\documentclass[11pt]{article}
\usepackage{enumerate}
\usepackage[OT1]{fontenc}
\usepackage[usenames]{color}
\usepackage{smile}
\usepackage[colorlinks,
            linkcolor=red,
            anchorcolor=blue,
            citecolor=blue
            ]{hyperref}
\usepackage{mathrsfs}
\usepackage{fullpage}
\usepackage{hyperref}
\usepackage[protrusion=true, expansion=true]{microtype}
\usepackage{float}
\usepackage{subfigure}
\usepackage{amsfonts,amsmath,amssymb,amsthm,url,xspace}
\usepackage{tikz}
\usepackage{verbatim}
\usetikzlibrary{arrows,shapes}
\usepackage{mathtools}
\usepackage{authblk}
\usepackage[bottom]{footmisc}
\usepackage{enumitem}
\usepackage{mathtools}
\usepackage{lscape}
\usepackage{caption}

\usepackage{algorithmicx,algpseudocode}

\usepackage{amsmath,amsfonts,bm}

\def\vtheta{{\bm{\theta}}}

\def\vs{{\bm{s}}}

\def\vu{{\bm{u}}}
\def\vv{{\bm{v}}}

\DeclareMathAlphabet{\mathsfit}{\encodingdefault}{\sfdefault}{m}{sl}
\SetMathAlphabet{\mathsfit}{bold}{\encodingdefault}{\sfdefault}{bx}{n}

\def\sR{{\mathbb{R}}}

\newcommand{\E}{\mathbb{E}}

\def\veta{{\bm{\eta}}}

\ifx\counterwithout\undefined\usepackage{chngcntr}\fi
\counterwithout{equation}{section}

\allowdisplaybreaks[1]

\newcommand{\xhdr}[1]{\vspace{0.1mm}\noindent{{\bf #1.}}}
\usepackage{natbib}
\usepackage{xcolor}

\begin{document}

\title{A Simplified Framework for Contrastive Learning for Node Representations}

\author[1,2]{Ilgee Hong}
\author[1,3]{Huy Tran}
\author[1,4]{Claire Donnat}
\affil[1]{Department of Statistics, The University of Chicago}
\affil[2,3,4]{\textit {\{ilgee,huydtran,cdonnat\}@uchicago.edu}}
\date{}

\maketitle

\begin{abstract}
Contrastive learning has recently established itself as a powerful self-supervised learning framework for extracting rich and versatile data representations. Broadly speaking, contrastive learning relies on a data augmentation scheme to generate two versions of the input data and learns low-dimensional representations by maximizing a normalized temperature-scaled cross entropy loss (NT-Xent) to identify augmented samples corresponding to the same original entity \citep{chen2020SimCLR}. In this paper, we investigate the potential of deploying contrastive learning in combination with Graph Neural Networks \citep{scarselli2008graph} for embedding nodes in a graph. Specifically, we show that the quality of the resulting embeddings and training time can be significantly improved by a simple column-wise postprocessing of the embedding matrix, instead of the row-wise postprocessing via multilayer perceptrons (MLPs) that is adopted by the majority of peer methods. This modification yields improvements in downstream classification tasks of up to 1.5\% and even beats existing state-of-the-art approaches on 6 out of 8 different benchmarks. We justify our choices of postprocessing by revisiting the ``alignment vs. uniformity paradigm'' of \cite{wang2020understanding}, and show that column-wise post-processing improves both ``alignment" and ``uniformity'' of the embeddings.
\end{abstract}

\section{Introduction}\label{sec:1}

\begin{table}[ht]
	\caption{Architecture Comparisons between Self-supervised Node Representation Learning Methods}\label{tab:tab1_architecture}
	\begin{center}
		\begin{small}
			\begin{sc}
				\begin{tabular}{ l c c c c c c }
					\toprule
					Method & Target & MI-Estimator & Proj/Pred head & Asymmetric & Negative samples\\ 
					\midrule
					DGI & N-G & \checkmark & - & - & \checkmark \\
					MVGRL & N-G & \checkmark & \checkmark & \checkmark & \checkmark \\
					BGRL & N-N & - & \checkmark & \checkmark  & -  \\ 
					CCA-SSG & F-F & - & - & -  & -  \\ 
					GRACE & N-N & - & \checkmark & - & \checkmark \\
					\textbf{CLNR (Ours)} & N-N & - & - & - & \checkmark \\
					\bottomrule
				\end{tabular}
			\end{sc}
		\end{small}
	\end{center}
	\caption*{Table~\ref{tab:tab1_architecture} is extension of~\cite{zhang2021CCA-SSG}. \textit{Target}: comparison pair; N/G/F denotes node/graph/feature respectively. \textit{MI-Estimator}: parameterized mutual information estimator. \textit{Proj/Pred head}: parameterized projection or prediction head after GNN encoder. \textit{Asymmetric}: asymmetric architectures such as two separate encoders for two distinct augmentations or EMA and Stop-Gradient. \textit{Negative samples}: requiring negative samples when computing the loss.} 
\end{table}

Contrastive learning has become one of the most popular self-supervised methods for learning rich and versatile data representations ---  a success largely owing to its impressive performance across domains ranging from image classification \citep{chen2020SimCLR,grill2020BYOL,he2020MOCO,hjelm2018DIM,oord2018CPC,tian2020CMC} to sequence modeling \citep{kong2019mutual, oord2018representation,wu2020clear} and adversarial learning \citep{ho2020contrastive,grill2020BYOL}. By and large, contrastive learning's general outline consists in generating two (or more) perturbed versions of the data and in subsequently learning to recognize representations from  ``positive pair'' (i.e. embeddings corresponding to the same original entity) and to embed them close to one another in the embedding space. Conversely, ``negative pairs'' (i.e. embeddings corresponding to different original entity) are encouraged to lie farther apart. Remarkably, this simple ``model-free'' learning of differences \citep{gutmann2022statistical} between data views has been shown to achieve state-of-the-art performance in many downstream supervised tasks when combined with a simple linear classifier~\citep{chen2020SimCLR}. This result is quite surprising as the method performs well not only relative to other unsupervised embedding algorithms, but also to supervised end-to-end methods whose primary aim is to learn embeddings tailored to a particular task.
\\
\\
\xhdr{Contrastive Learning for Graphs}\hspace{11pt}Inspired by its success across various machine learning applications, contrastive learning has more recently been adopted by the graph representation learning community. In this case, contrastive learning is usually coupled with Graph Neural Networks (GNNs) \citep{scarselli2008graph, kipf2016semi,zhou2020graph,wu2020comprehensive} to learn a new class of node embeddings. Current contrastive methods for graphs can be classified into two categories, depending on which mutual information seeks to be maximized.
\\
\\
\textbf{1. Local-to-global mutual information maximization, } such as those used by Deep Graph InfoMax (DGI)~\citep{velickovic2019DGI} and Multi-View Graph Representation Learning (MVGRL)~\citep{hassani2020mvgrl}. These methods seek to learn node embeddings by maximizing the mutual information between representations of the original data at different scales: DGI considers the mutual information between local (node) representations obtained by passing the original graph through a GNN encoder and a global graph embedding obtained by summing all local representations. Negative samples are generated by a corruption function separately. MVGRL, on the other hand, leverages diffusion processes to generate additional graphs and obtain corresponding additional pairs of node and graph representations. A pair of node and graph representations from different graph are considered as negative pairs. Both DGI and MVGRL use mutual information estimator parameterized by a neural network to obtain probability score for each local and global representation pair. More specifically, letting $G$ and $\widetilde{G}$ denote the original and the corrupted graphs respectively, the DGI loss writes as: 
\begin{align*}
	\mathcal{L} = \frac{1}{N+M}\left(
	\sum_{i=1}^N
	\mathbb{E}_{G}
	\left[
	\log \mathcal{D}\left(\vu_i, \vs\right)
	\right] + 
\sum_{j=1}^M\mathbb{E}_{\widetilde{G}}
\left[
\log\left(1 - \mathcal{D}(\widetilde{\vu}_j,\vs)\right)
\right]
\right),
\end{align*}
where $\mathcal{D}:\sR^{F}\times\sR^{F}\rightarrow\sR$ is a mutual information estimator parameterized by MLPs, $\vs\in\sR^F$ is the summary vector of the original graph, and $\vu_i$, $\widetilde{\vu}_j\in\sR^F$ are node representations from the original graph and its corrupted version respectively with $i=1,...,N$ and $j=1,...,M$. 
\\
\\
\textbf{2. Local-to-local mutual information maximization,} such as Graph Contrastive Representation Learning (GRACE)~\citep{zhu2020GRACE} and its extension, Graph Contrastive representation learning with Adaptive augmentation (GCA)~\citep{zhu2021GCA}. Contrary to the previous approaches, these methods directly define similarity scores using the dot product between node pairs and foregoes the need for a parameterized mutual information estimator. These methods typically project their node representations onto the unit hypersphere, which is known to improve training stability and is more easily separable by downstream linear classifiers. GRACE and GCA generate two views of the original graph by randomly perturbing both edges and features (typically through a combination of random edge deletions and feature masking). Note that perturbations here are used to generate alternative versions of the same data rather than negative examples. Node representations are then learned through the maximization of a normalized temperature-scaled cross entropy loss (NT-Xent)~\citep{chen2020SimCLR} between node pairs, which amounts to solving a classification problem: for each node, in each version of the data, the algorithm seeks to recognize the embeddings from the same original node amongst all other embeddings in both versions of the graph. The objective function is provided in Equations~\eqref{eq:main_loss1} and~\eqref{eq:main_loss2}. Despite the impressive performance showed by GRACE and GCA, they suffer from a few drawbacks. Both methods rely on row-wise postprocessing of a raw embedding matrix via a MLP with a nonlinear activation function. This choice was first inspired by the method's vision counterpart \citep{chen2020SimCLR} where it was empirically shown that utilizing a 2-layers MLPs with a nonlinear activation function after the ResNet~\citep{he2016resnet} improved the expressive power and quality of representations compared to those with a linear activation function or without postprocessing step. However, the use of additional postprocessing via MLPs considerably complexifies the model architecture and its effect on the resulting representations is still ill-understood. 
\\
\\
\xhdr{Non-contrastive Learning for Graphs} The previous set of methods have some computational difficulties when processing large graph as the evaluation of the normalizing factor for each node requires the computation of pairwise cosine similarities between all negative pairs. Consequently, newer methods such as Bootstrapped Graph Latents (BGRL)~\citep{thakoor2021BGRL} and Canonical Correlation Analysis inspired Self-Supervised Learning on Graphs (CCA-SSG)~\citep{zhang2021CCA-SSG} propose to leverage non-contrastive objectives which do not incorporate the negative samples. In CCA-SSG, for instance, the goal is to directly minimize the distance between positive pairs, while additionally decorrelating features in different dimensions. The objective function writes as:
\begin{equation*}
	\mathcal{L}=\left\Vert U-V\right\Vert_F^2+\lambda\left(\left\Vert U^T U- I\right\Vert_F^2+\left\Vert V^T V-I\right\Vert_F^2\right)
\end{equation*}
where $U, V\in\sR^{N\times F}$ are node representation matrix of two perturbed versions of the original graph, $\lambda>0$ be the hyperparameter, and $\Vert\cdot\Vert_F$ denotes the Frobenius norm. In consequences, both BGRL and CCA-SSG enjoy low computational complexities by disregarding any computations with negative samples. However, BGRL requires additional components such as exponential moving average (EMA) and Stop-Gradient --- making the method both more expensive and harder to interpret. Although CCA-SSG relies on a simple and symmetric architecture without additional components, this feature-wise contrastive method typically requires larger embedding dimension to improve the quality of its representations.
\\
\\
\xhdr{Contributions} The contributions of our paper are threefold. (1) We propose a simplified framework for contrastive learning for node representations and refer to our method as \textbf{\textit{Contrastive Learning for Node Representations (CLNR)}}. CLNR follows the core structure of GRACE~\citep{zhu2020GRACE} and GCA~\citep{zhu2021GCA} which generate two views of an input graph via random data augmentation, use a shared GNN encoder, and find node representations by maximizing the NT-Xent loss~\citep{chen2020SimCLR}. In contrast to GRACE and GCA, CLNR simplifies the model architecture by replacing row-wise postprocessing of a node representation matrix via a MLPs with a simple column-wise standardization. We show this substitution considerably improves the quality of representations and model training time. (2) We provide a finer understanding on the effect of various postprocessing steps and justify our choice of postprocessing (column-wise standardization) by revisiting the ``alignment and uniformity paradigm'' of \cite{wang2020understanding} which was proposed to analyze the quality of representations on the unit hypersphere. (3) We provide experimental evidence of the quality of the induced node embeddings by using them as input for downstream classification. We show that CLNR provides superior results compared to state-of-the-art approaches, including CCA-SSG and GRACE. 
\\
\\
\xhdr{Notations} For a given graph $G=(X,A)$, $X\in\sR^{N\times D}$ is the feature matrix and $A\in\{0,1\}^{N\times N}$ is the adjacency matrix. Here, $N$ denotes the number of nodes in the graph $G$, while $D$ denotes the feature dimension. 
\\
\\
\xhdr{Structure of the paper} We introduce the general contrastive learning framework for node representations and propose our simplified framework in Section~\ref{sec:sec2}. Experiments are given in Section~\ref{sec:sec2-2} and discussion on various postprocessing step is provided in Section~\ref{sec:sec3}. Additional experiments are present in Section~\ref{sec:sec4} and conclusions are presented in Section~\ref{sec:sec5}.

\section{Contrastive Learning for Node Representations}\label{sec:sec2}

We begin by presenting the general framework for contrastive learning for node embddings, and focus on studying the choice of the embedding postprocessing.

\subsection{Model Framework}

\xhdr{Graph Augmentations} We follow the standard pipeline for random graph augmentation developed by~\cite{zhu2020GRACE}, which comprises two components: {\it (1) edge dropping} and {\it (2) feature masking}. This procedure thus relies on two distinct hyperparameters: the ``edge drop rate" (that is, the Bernoulli probability for each edge to  be dropped, independently from all the others), and the ``feature mask rate'' (or probability that each element of  the feature matrix to be masked, independently from the others). This set of perturbations allow the generation of two alternative versions of the same input graph $G$ at each epoch. Let us denote these two versions as $G_1$ with $(X_1, A_1)$ and $G_2$ with $(X_2,A_2)$. 
\\
\\
\xhdr{Shared GNN Encoder} We embed the nodes in both $G_1$ and $G_2$ in Euclidean space using a shared Graph Neural Network (GNN) \citep{scarselli2008graph}. Let $f_{\vtheta}$ be the shared GNN encoder. We denote the raw embedding matrices by $f_{\vtheta}(X_1,A_1)=Z_1$ and $f_{\vtheta}(X_2,A_2)=Z_2$ where $Z_1, Z_2\in
\sR^{N\times F}$ and $F$ denotes the embedding dimension. 
\\
\\
\xhdr{Postprocessing Step} The outcome of the previous convolution step is a set of raw embedding matrices $Z_1$ and $Z_2$ for the two versions of the input graph. Empirical evidence shows that further postprocessing of these raw embeddings can greatly improve their expressive power. Usual options include: 

\begin{description}[topsep=0em, itemsep=0em,leftmargin=0em]
\item [(1) Row-wise Postprocessing,] in which case each node embedding is processed independently of the others. Let $g_{\veta}$ be a 2-layers MLPs with a nonlinear activation function. The final node embedding matrices for each of the two versions of the data can be written as $g_{\veta}(Z_1)=U$ and $g_{\veta}(Z_2)=V$. Such is the setting adopted for instance by GRACE~\citep{zhu2020GRACE}, which closely follows the architecture of SimCLR~\citep{chen2020SimCLR}. The latter showed that adding a 2-layers MLPs with a non-linear activation function to the ResNet~\citep{he2016resnet} outperformed other choices such as MLPs with a linear activation function or the one without postprocessing for vision tasks~\citep{chen2020SimCLR,bachman2019lrmi}. 
\item [(2) Column-wise Postprocessing] --- and more specifically, column-wise standardization, in which case each dimension of the raw embeddings is centered and rescaled. Therefore, each node embedding is processed along with others in a batch, and no longer independently. This type of processing mechanism is referred to as ``batch normalization'' (BN) in the deep learning literature~\citep{ioffe2015batch, santurkar2018does, bjorck2018understanding}. When training deep neural networks (DNNs), BN is typically used before the activation function between layers to standardize inputs. BN was initially proposed to alleviate internal covariate shift, defined as the shift in the distribution of network activations caused by the change in network parameters during training~\citep{ioffe2015batch}. BN was subsequently found to improve speed and stability of the model's training and generalization accuracy by smoothing the optimization landscape~\citep{santurkar2018does} and allowing faster learning rates~\citep{bjorck2018understanding}. Here, we propose a new way of using BN as a postprocessing step in self-supervised contrastive learning. {\it We posit indeed that the row-wise MLPs postprocessing used in usual methods overly complexifies the architecture of the algorithm with no clear gain in performance.} Rather, in Sections~\ref{sec:sec2-2} and~\ref{sec:sec3}, we will show that the column-wise BN postprocessing results in better representations and this result is supported by recent findings in the contrastive learning literature of alignment and uniformity paradigm~\citep{wang2020understanding}. We refer to the contrastive learning framework with this choice of postprocessing as {\textit{Contrastive Learning method for Node Representations (CLNR)}}. The final node embeddings can be written as follows:
	\begin{equation}
		U=\frac{Z_1-\text{mean}(Z_1)}{\text{std}(Z_1)}\quad\text{and}\quad V=\frac{Z_2-\text{mean}(Z_2)}{\text{std}(Z_2)},
	\end{equation}
	where for any matrix $A\in\sR^{n\times q}$, $\text{mean}(A)\in\sR^{n\times q}$ and $[\text{mean}(A)]_{i,j}=\sum_{k=1}^n a_{k,j}/n$ and $\text{std}(A)\in\sR^{n\times q}$ and $[\text{std}(A)]_{i,j}=\sqrt{\sum_{k=1}^n(a_{k,j}-[\text{mean}(A)]_{i,j})/n}$.\\
\end{description}
	
\xhdr{Projection onto Hypersphere} Regardless of the choice of postprocessing, the embeddings are then finally projected onto the unit hypersphere. This final step stems from the substantial deep learning literature \citep{bojanowski2017unsupervised,chen2020SimCLR,davidson2018hyperspherical,meng2019spherical,xu2018spherical}. This $\ell_2$ normalization of the final embedding matrices (e.g., $U$ and $V$) is indeed known to improve the stability of learning process \citep{wang2020understanding,xu2018spherical}, and makes the embeddings more easily separable by downstream linear classifiers.
\\
\\
\xhdr{Contrastive Loss Function} 
We use the contrastive objective from~\cite{zhu2020GRACE}, which is called ``NT-Xent loss" where the term was first used in~\cite{chen2020SimCLR}. Contrary to GRACE --- which computes the cosine similarities of all pairs of nodes in the graph ---, we randomly subsample $m$ nodes at each epoch and only compute the cosine similarity between these nodes. This allows the method to scale well for a large graph with no expenses in accuracy if $m$ is sufficiently large. For each node $i=1,...,m$ in the random mini-batch, we let $\vu_i=U_{i,:}$ and $\vv_i=V_{i,:}$ be the $i$-th row of $U$ and $V$ respectively. Then, for each $i=1,...,m$, if $\vu_i$ is the target node, $(\vu_i,\vv_i)$ is viewed as a positive pair, and $(\vu_i,\vv_k)$ and $(\vu_i,\vu_k)$ where $k\neq i$ are naturally considered as negative pairs. Note that the first negative pairs are from inter-view nodes, and the second negative pairs are from intra-view nodes. Let $s:\sR^F\times \sR^F\rightarrow\sR$ be the cosine similarity function (defined as $s(\vu,\vv)=\vu\cdot \vv/\Vert \vu\Vert\Vert \vv\Vert$).  Our objective function takes the following explicit form:
\begin{align}\label{eq:main_loss1}
\ell(\vu_i,\vv_i)=\log\frac{\overbrace{e^{s(\vu_i,\vv_i)/\tau}}^{\textit{positive pair}}}{\underbrace{e^{s(\vu_i,\vv_i)/\tau}}_{\textit{positive pair}}+\underbrace{\sum_{k=1}^{m}\bm{1}_\mathrm{[k\neq i]}\left[e^{s(\vu_i,\vu_k)/\tau}+e^{s(\vu_i,\vv_k)/\tau}\right]}_{\textit{negative pairs}}},
\end{align}
where $\tau$ denotes the temperature parameter. It is worth noting that in general, $\ell(\vu_i,\vv_i)\neq\ell(\vv_i,\vu_i)$. Therefore, taking the average over all possible pairs, the final objective to be maximized is given by
\begin{equation}\label{eq:main_loss2}
	\mathcal{L} = \frac{1}{2m}\left(\sum_{i=1}^m\left[\ell(\vu_i,\vv_i)+\ell(\vv_i,\vu_i)\right]\right).    
\end{equation}

\section{Experiments} \label{sec:sec2-2}

In this section, we showcase the potential of our simplified framework for node constrative learning by deploying it on a number real-life datasets and comparing it against existing peer methods.
\\
\\
\xhdr{Real-life Datasets} We evaluate the quality of our node representations using a \textbf{node classification task} on real-life data. We consider eight node classification benchmarks: \textit{Cora, CiteSeer, Pubmed, Amazon Computers, Amazon Photo, Coauthor CS, Coauthor Physics, ogbn-arxiv}. The first seven benchmarks are used in \cite{Yang2019Revisiting} and \cite{shchur2018pitfalls}, and \textit{ogbn-arxiv} is from the Open Graph Benchmark datasets \citep{hu2020OGB}. We adopt the public splits for the citation network datasets (\textit{Cora, Citeseer, Pubmed}), and 1:1:8 train/validation/test splits for Amazon and Coauthor network datasets (\textit{Amazon Computer, Amazon Photo, Coauthor CS, Coauthor Physics}). Details of the datasets are given in Appendix~\ref{appen:A1}.
\\
\\
\xhdr{Experimental Setup} We follow the evaluation protocol suggested by \cite{velickovic2019DGI}, which consists of two stages: 1) Pretraining and  2) Linear evaluation. First, we begin by training a GNN encoder using the NT-Xent loss described in Equation~\eqref{eq:main_loss1}, thereby yielding node embeddings for the entire graph. We then fit a simple linear classifier on a subset of nodes (training set). We then evaluate the classification accuracy on the remaining nodes  (test set). All experiments are done with Pytorch and Pytorch Geometric module on GPU with 16 GB memory.
\begin{table}[t]
	\caption{\label{tab:tab4_accuracy} Node Classification Accuracy}
	\begin{center}
		\begin{small}
			\begin{sc}
				\resizebox{\textwidth}{!}{
					\begin{tabular}{lccccccccc} 
						\toprule
						Methods & Input & Cora & CiteSeer & PubMed & Computers & Photo & CS & Physics & ogbn-arxiv\\
						\midrule
						GAE & X, A & 71.5$\pm$0.4 & 65.8$\pm$0.4  & 72.1$\pm$0.5 & 85.3$\pm$0.2 & 91.6$\pm$0.1  & 90.0$\pm$0.7 & 94.9$\pm$0.1 & 52.57$\pm$0.1 \\ 
						
						DGI & X, A &82.3$\pm$0.6 & 71.8$\pm$0.7  & 76.8$\pm$0.6 & 84.0$\pm$0.5 & 91.6$\pm$0.2  & 92.2$\pm$0.6 & 94.5$\pm$0.5 & 67.95$\pm$0.6\\
						
						MVGRL & X, A, S & 83.5$\pm$0.4 & \textbf{73.3$\pm$0.5} & 80.1$\pm$0.7 & 87.5$\pm$0.1 & 91.7$\pm$0.1 & 92.1$\pm$0.1 & 95.3$\pm$0.1 & OOM\\
						
						CCA-SSG & X, A & 84.2$\pm$0.4 & 73.1$\pm$0.3  & 81.6$\pm$0.4  & 88.7$\pm$0.3 & 93.1$\pm$0.1  & 93.3$\pm$0.2 & 95.4$\pm$0.1 &  60.5 $\pm$ 0.2\\ 
						
						BGRL & X, A & 82.1$\pm$1.0 & 71.8$\pm$0.5 & 80.6$\pm$1.0 & 89.0$\pm$0.4 &  92.7$\pm$0.4 & 92.0$\pm$0.2 & 94.5$\pm$0.3 & 70.3$\pm$0.1 \\ 
						
						GRACE & X, A & 83.6$\pm$0.8 & 71.8$\pm$1.1 & 81.6$\pm$0.9 &89.1$\pm$0.4 & \textbf{93.3$\pm$0.4} & 93.6$\pm$0.3 & 95.3$\pm$0.1 &  69.8 $\pm$ 0.1\\
						
						\textbf{CLNR} & X, A & \textbf{84.3$\pm$0.6} & 73.1$\pm$0.7  & \textbf{83.0 $\pm$ 0.6} & \textbf{89.5$\pm$0.4} &  93.2$\pm$0.3 & \textbf{93.8$\pm$0.2} &  \textbf{95.5$\pm$0.1} & \textbf{70.4 $\pm$ 0.3} \\ 
						\midrule						
						GCN &  X, A, Y & 81.5$\pm$0.7 & 70.3$\pm$0.7 & 79.0$\pm$0.3 & 86.5$\pm$0.5 & 92.4$\pm$0.2 & 93.0$\pm$0.3 & 95.7$\pm$0.2 & 71.7$\pm$0.3  \\
						
						GAT & X, A, Y & 83.0$\pm$0.7 & 72.5$\pm$0.7  & 79.0$\pm$0.3 & 86.9$\pm$0.3 & 92.6$\pm$0.4  & 92.3$\pm$0.2 & 95.5$\pm$0.2 & 71.7$\pm$0.3\\
						\bottomrule
					\end{tabular}
				}
			\end{sc}
		\end{small}
	\end{center}
	\caption*{Mean classification accuracy with standard deviation for node classification task on 7 benchmarks. \textit{Input}:  $X$: node features, $A$: adjacency matrix, $S$: diffusion matrix, and $Y$: node labels. For GRACE and CLNR, we use $m=1024$ in the loss shown in Equation~\eqref{eq:main_loss1}. OOM indicates out-of-memory with 16GB GPU.}
\end{table}
\\
\\
\xhdr{Implementation Details} For this first set of experiments, we took our graph encoder $f_{\vtheta}$ to be a standard 2-layers GCN model~\citep{kipf2016semi} with the same hidden and output dimensions across all datasets except \textit{CiteSeer}, \textit{Coauthor CS}, and \textit{ogbn-arxiv}. As in \citet{zhang2021CCA-SSG}, we observed 1-layer GCN is better for \textit{CiteSeer} and 2-layers MLPs is better for \textit{Coautor CS} than using 2-layers GCN. We use a 3-layers GCN for \textit{ogbn-arxiv} due to the difficulty of the task (40 classes are in \textit{ogbn-arxiv}, compared to less than 15 classes in the others). For the second stage, we use a simple logistic regression as a linear classifier. We utilize the Adam optimizer \citep{kingma2014adam}. We run 20 random initializations on both model parameters and data splits, and report the mean accuracy along with a standard deviation. Additional details regarding hyperparameter tuning are provided in Appendix~\ref{appen:A2}.
\\
\\
\xhdr{Results} For fair comparison, we provide a consistent training setting across all methods: we use an Adam optimizer and a simple logistic regression without $\ell_2$ regularization for all methods and for all benchmark datasets. Table~\ref{tab:tab4_accuracy} provides the comparison between CLNR and existing peer methods by reporting the mean classification accuracy (and standard deviation). We first focus on comparisons between CLNR (column-wise postprocessing) and GRACE (row-wise postprocessing). We can see that CLNR outperforms GRACE up to 1.5\% throughout all benchmark datasets --- with the exception of \textit{Photo}. This shows that column-wise postprocessing achieves better representations than row-wise postprocessing in general. We also compare the performance of CLNR against other existing methods. In particular, we compare CLNR with one classical unsupervised model: GAE~\citep{kipf2016vgae}, five self-supervised models: DGI~\citep{velickovic2019DGI}, MVGRL~\citep{hassani2020mvgrl}, BGRL~\citep{thakoor2021BGRL}, CCA-SSG~\citep{zhang2021CCA-SSG}, and two supervised baselines: GCN~\citep{kipf2016semi} and GAT~\citep{velivckovic2017gat}. From Table~\ref{tab:tab4_accuracy}, we observe that CLNR outperforms all other unsupervised and self-supervised methods on all benchmarks except \textit{CiteSeer}, where MVGRL is superior. CLNR even outperforms two supervised baselines, GCN and GAT, by up to 1 to 4\% on all benchmark datasets except \textit{Coauthor Physics} and \textit{ogbn-arxiv}. On the other hand, among methods using non-contrastive losses, CCA-SSG outperforms BGRL on all benchmark datasets except \textit{Computer} and \textit{ogbn-arxiv}.
\\
\\
\xhdr{Training Time} It is also worth emphasizing that CLNR typically needs smaller number of epochs to generate the best node representations. This is due to other advantage of batch normalization which allows a higher learning rate and make the optimization landscape smoother, hence, accelerating model training. Moreover, CLNR does not need to learn additional parameters in a MLPs. Table~\ref{tab:time_comp} shows the comparison between GRACE and CLNR in terms of model training time. We observe that CLNR is up to 20 times faster in training than GRACE.

\begin{table}[h]
	\caption{Training Time of CLNR and GRACE.} \label{tab:time_comp}
	\begin{center}
		\begin{small}
			\begin{sc}
				\begin{tabular}{lcc}
					\toprule
					Model & Dataset & Training time\\
					\midrule
					\textbf{CLNR} & ogbn-arxiv & \textbf{2879.75s} \\
					GRACE & ogbn-arxiv &  7286.71s \\
					\midrule
					\textbf{CLNR} & Photo & \textbf{7.87s} \\
					GRACE & Photo &  143.25s \\
				    \midrule					
					\textbf{CLNR} & Computers & \textbf{25.62s} \\
					GRACE & Computers &  104.59s \\
						\midrule					
					\textbf{CLNR} & PubMed & \textbf{134.51s} \\
					GRACE & PubMed &  538.75s \\
					\bottomrule
				\end{tabular}
			\end{sc}
		\end{small}
	\end{center}
\end{table}

\section{Revisiting the ``Alignment vs. Uniformity Paradigm"}\label{sec:sec3}

In this section,  we propose to delve deeper in the comparison between CLNR and GRACE, and try to explain the superiority of the embeddings obtainted by CLNR.
To this end, we revisit the ``alignment and uniformity paradigm'' proposed by \cite{wang2020understanding}.
\\
\\
\xhdr{Alignment vs. Uniformity} Noting the success of mapping features to the unit hypersphere and restricting their analysis to those, \cite{wang2020understanding} propose an analysis of contrastive losses in terms of two quantities: {\it (a) Alignment}, which quantifies how close representations from positive pairs (i.e. embeddings corresponding to the same entity) are mapped to one another: \begin{equation}\label{eq:align}
	\ell_{\text{align}} = \E_{(x,y)\sim p_{\text{pos}}} \| f(x) - f(y)\|_2^{\alpha}.
\end{equation} 
In the previous expression, $f(x)$ denotes the embedding for the nodes, $\alpha>0$ is any positive exponent (usually chosen to be equal to 2);  And {\it (b) uniformity}, defined as the logarithm of the average pairwise Gaussian potential: 
\begin{equation}\label{eq:unif}
	\ell_{\text{uniform}}(f,t)
	= \log \E_{ x, y \overset{\text{i.i.d}}{\sim} p_{\text{data}}} \Big[{e^{-t \|{f(x) - f(y)\|_2^2}}}\Big]
	, \quad t > 0.%
\end{equation}
Uniformity essentially measures uniformly distributed the features are on the unit sphere. \cite{wang2020understanding} indeed posit that data distributions that are more uniformly spread on the hypersphere preserve more information.  They then proceed to show that the NT-Xent loss can asymptotically be written as the sum of a uniformity and an alignment term, so that it optimizes for both. Building off of these observations, the authors suggest that explicitly optimizing for both uniformity and alignment might yield better results than using the classical negative cross-entropy loss. However, the weighted loss that they propose $\ell = \ell_{\text{uniform}}  +    \lambda \ell_{\text{align}}   $ has the disadvantage of introducing  yet another parameter $\lambda$ that has to be cross-validated for.
 \begin{figure}[t]
	\begin{minipage}[h]{.19\linewidth}
		\centerline{\includegraphics[scale=0.142]{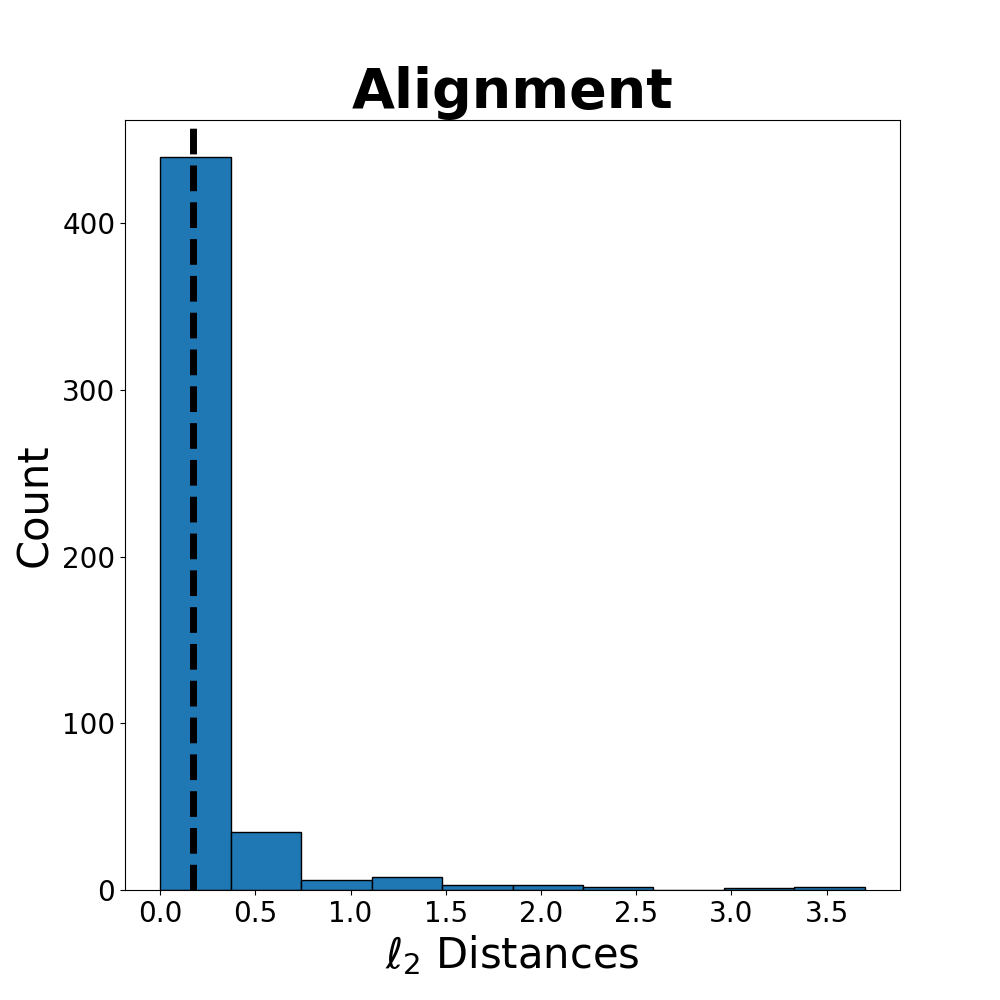}}
	\end{minipage}
	\begin{minipage}[h]{.19\linewidth}
		\centerline{\includegraphics[scale=0.142]{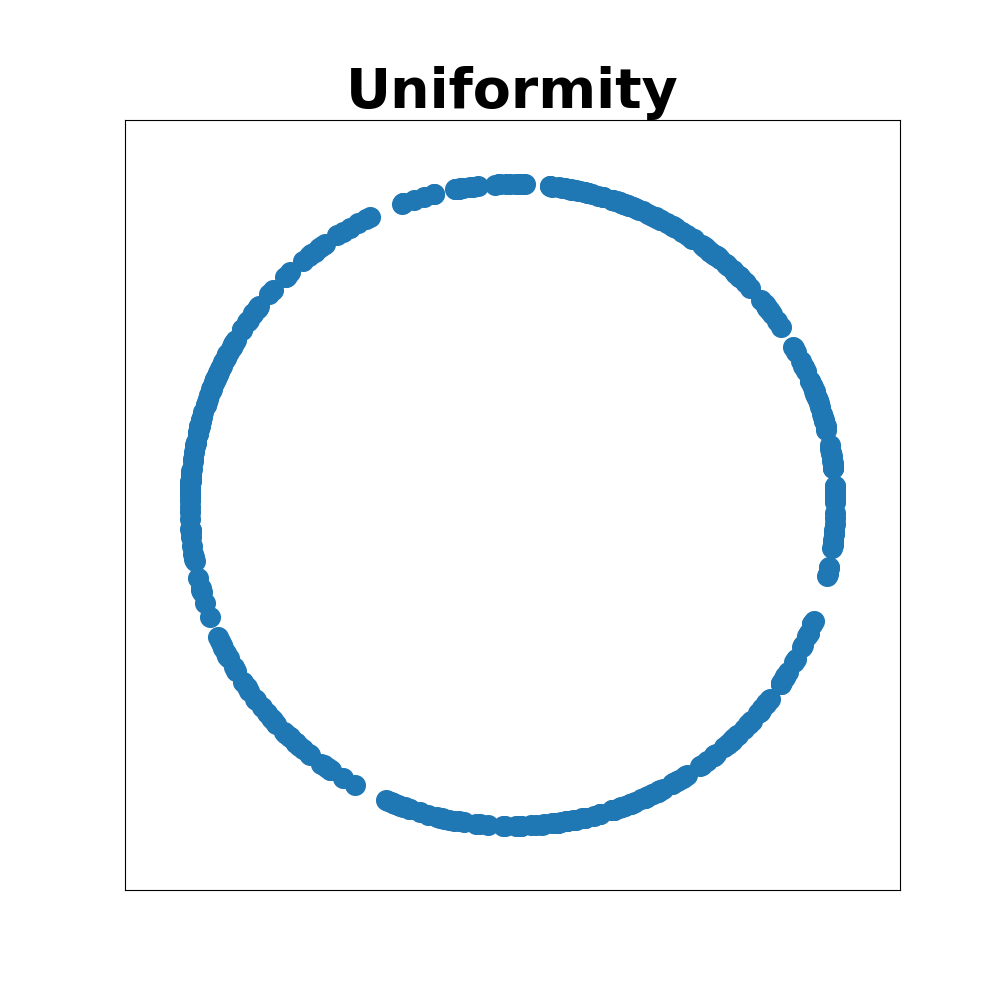}}
	\end{minipage}
	\begin{minipage}[h]{.19\linewidth}
		\centerline{\includegraphics[scale=0.142]{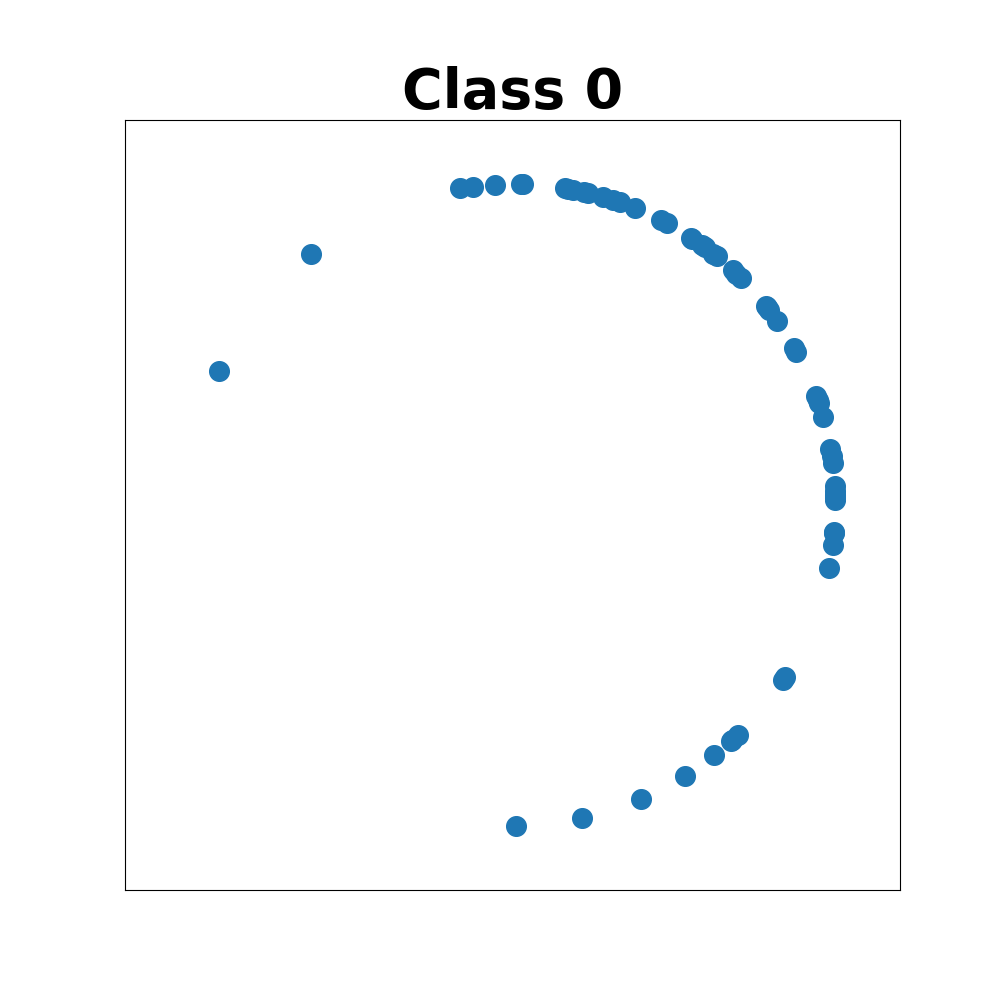}}
	\end{minipage}
	\begin{minipage}[h]{.19\linewidth}
		\centerline{\includegraphics[scale=0.142]{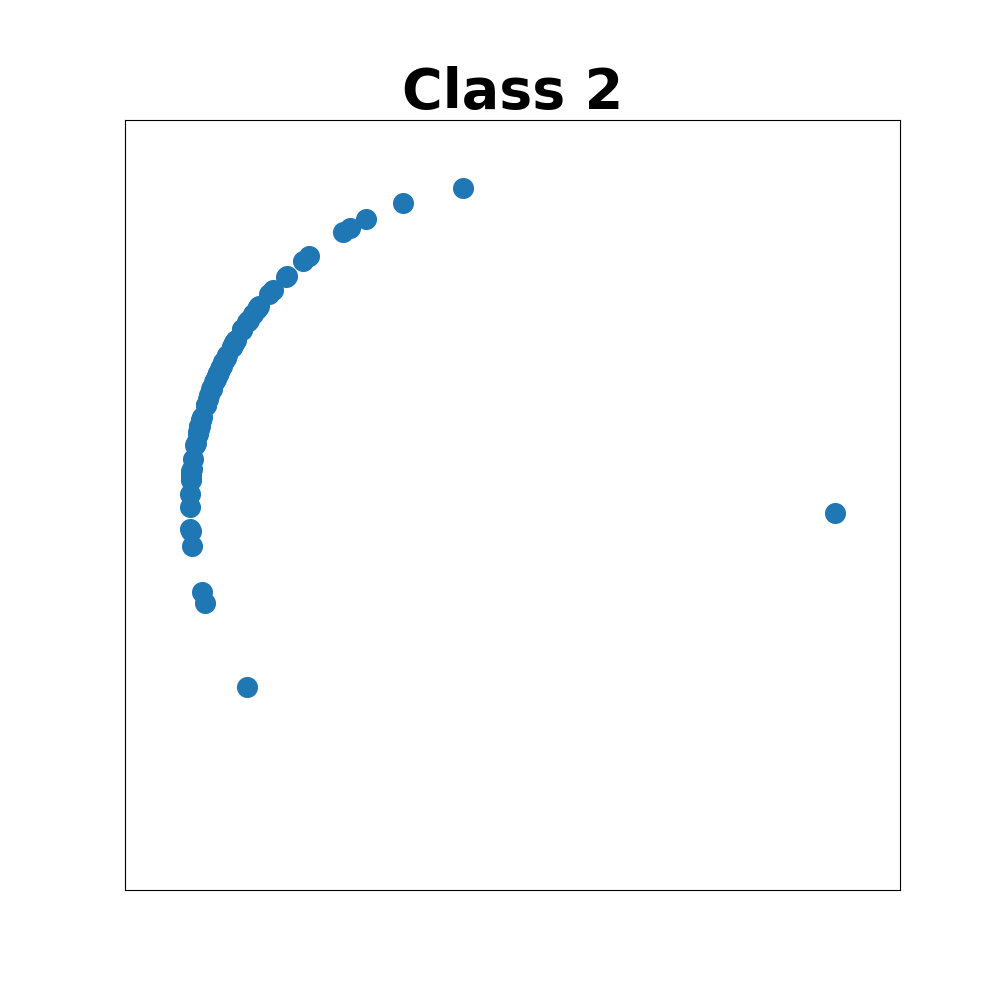}}
	\end{minipage}
	\begin{minipage}[h]{.19\linewidth}
		\centerline{\includegraphics[scale=0.142]{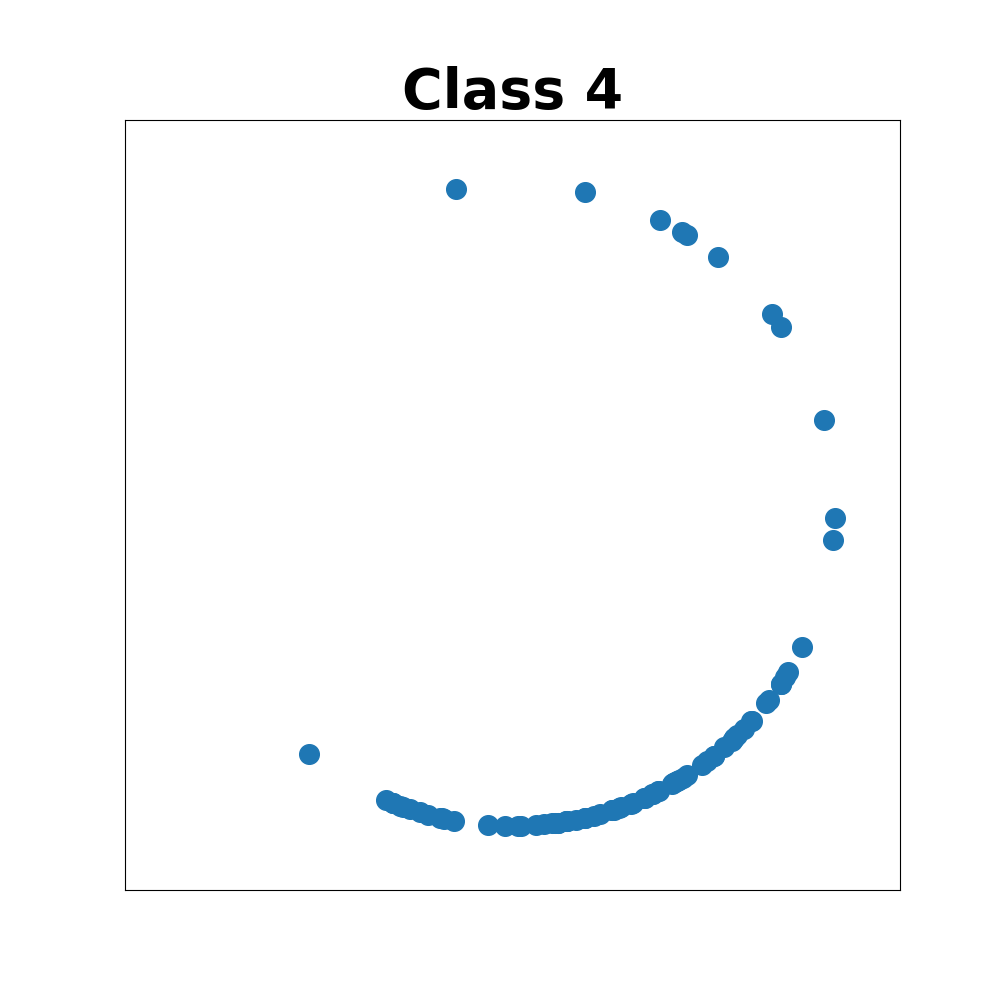}}
	\end{minipage}
	\\
	\begin{minipage}[h]{.19\linewidth}
		\centerline{\includegraphics[scale=0.142]{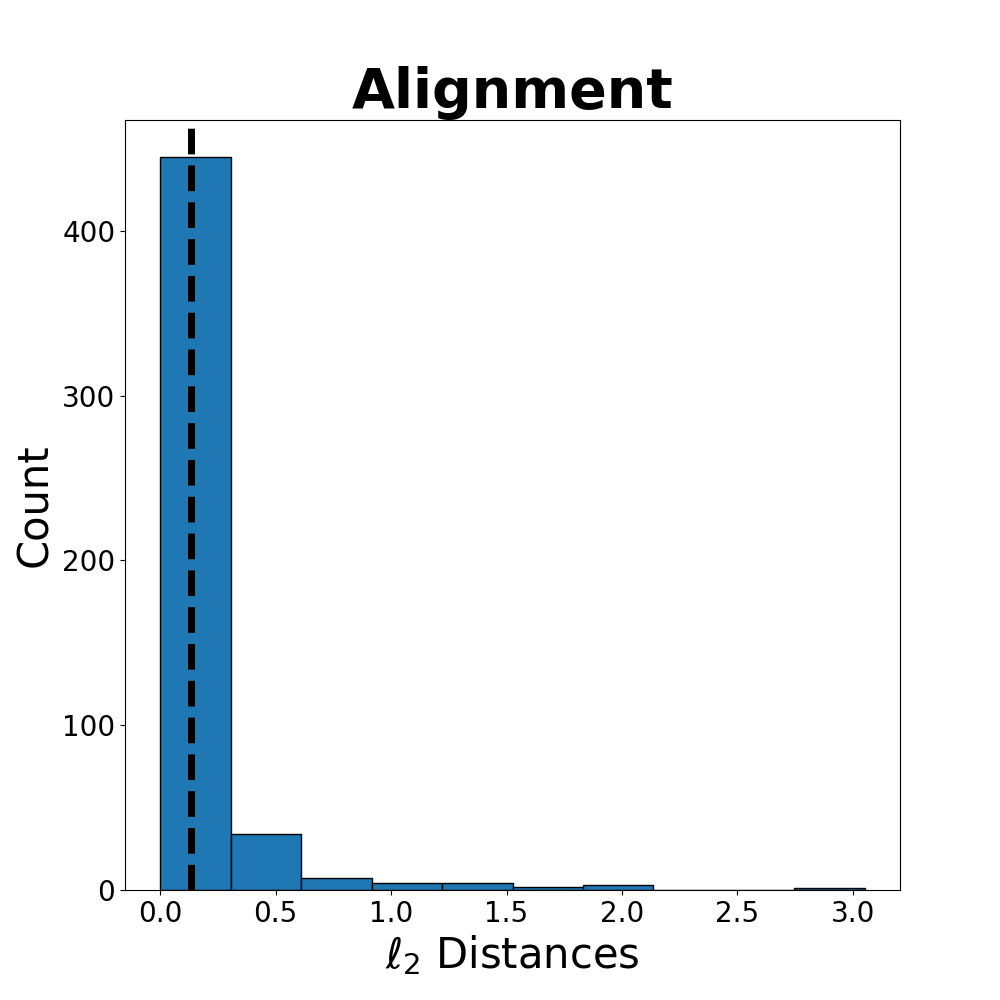}}
	\end{minipage}
	\begin{minipage}[h]{.19\linewidth}
		\centerline{\includegraphics[scale=0.142]{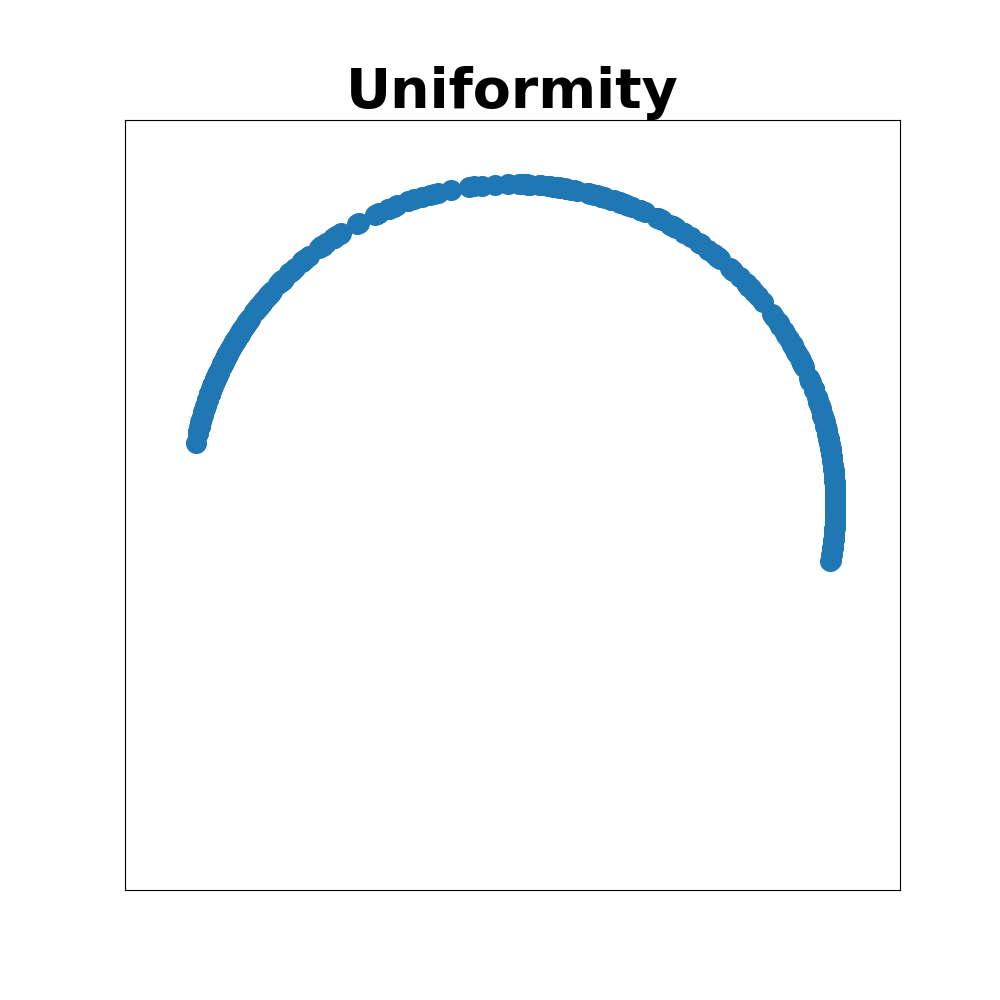}}
	\end{minipage}
	\begin{minipage}[h]{.19\linewidth}
		\centerline{\includegraphics[scale=0.142]{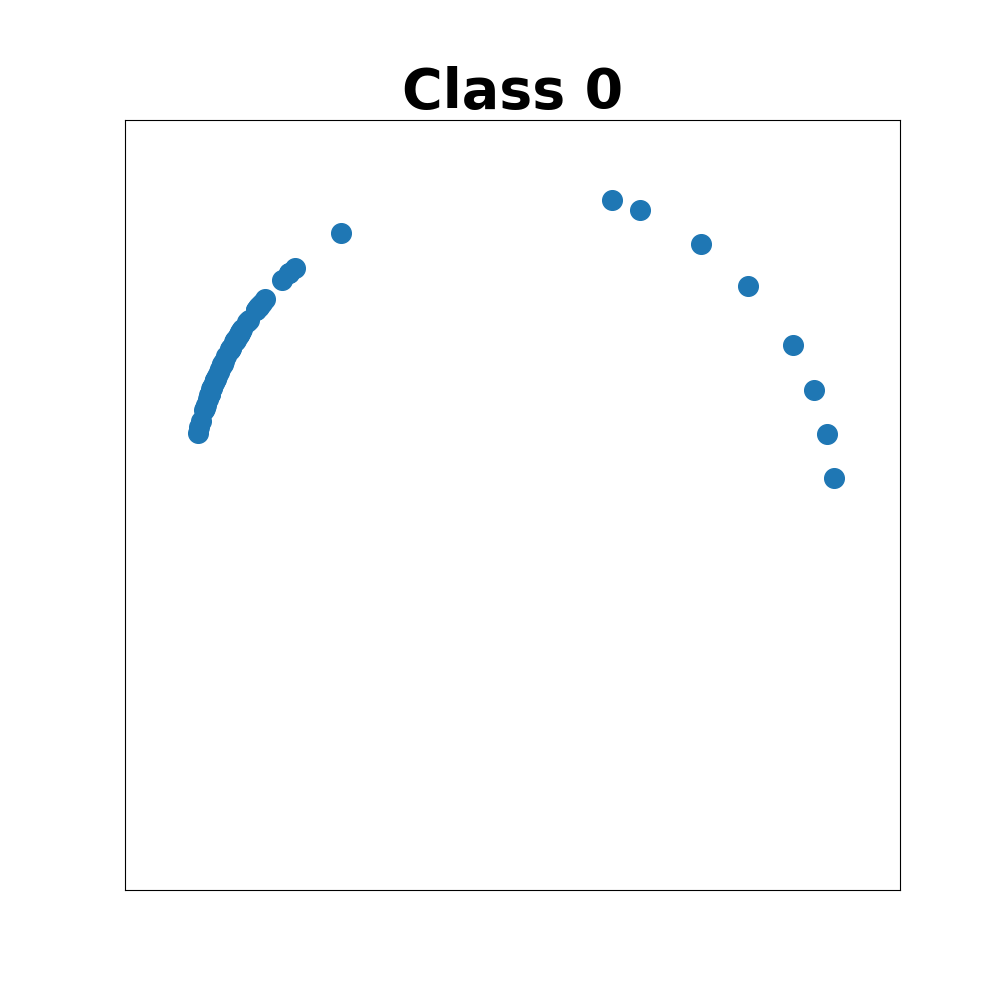}}
	\end{minipage}
	\begin{minipage}[h]{.19\linewidth}
		\centerline{\includegraphics[scale=0.142]{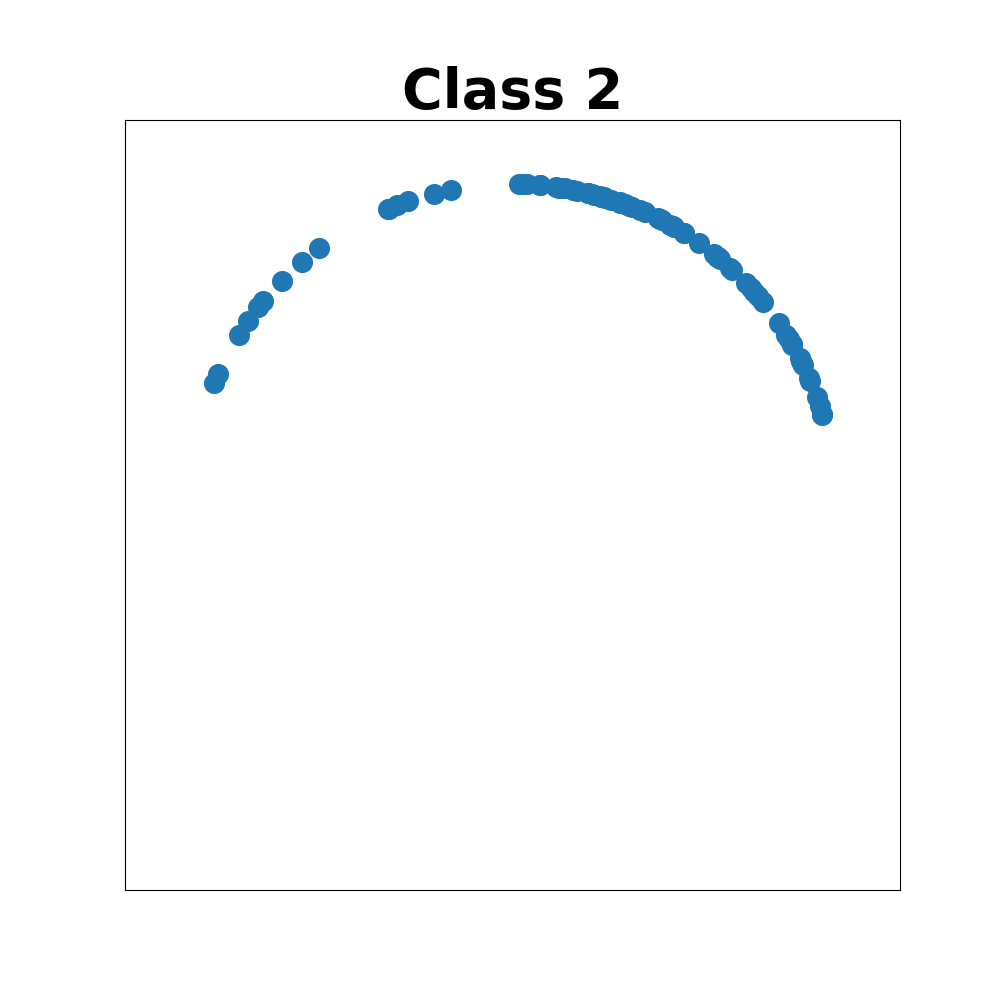}}
	\end{minipage}
	\begin{minipage}[h]{.19\linewidth}
		\centerline{\includegraphics[scale=0.142]{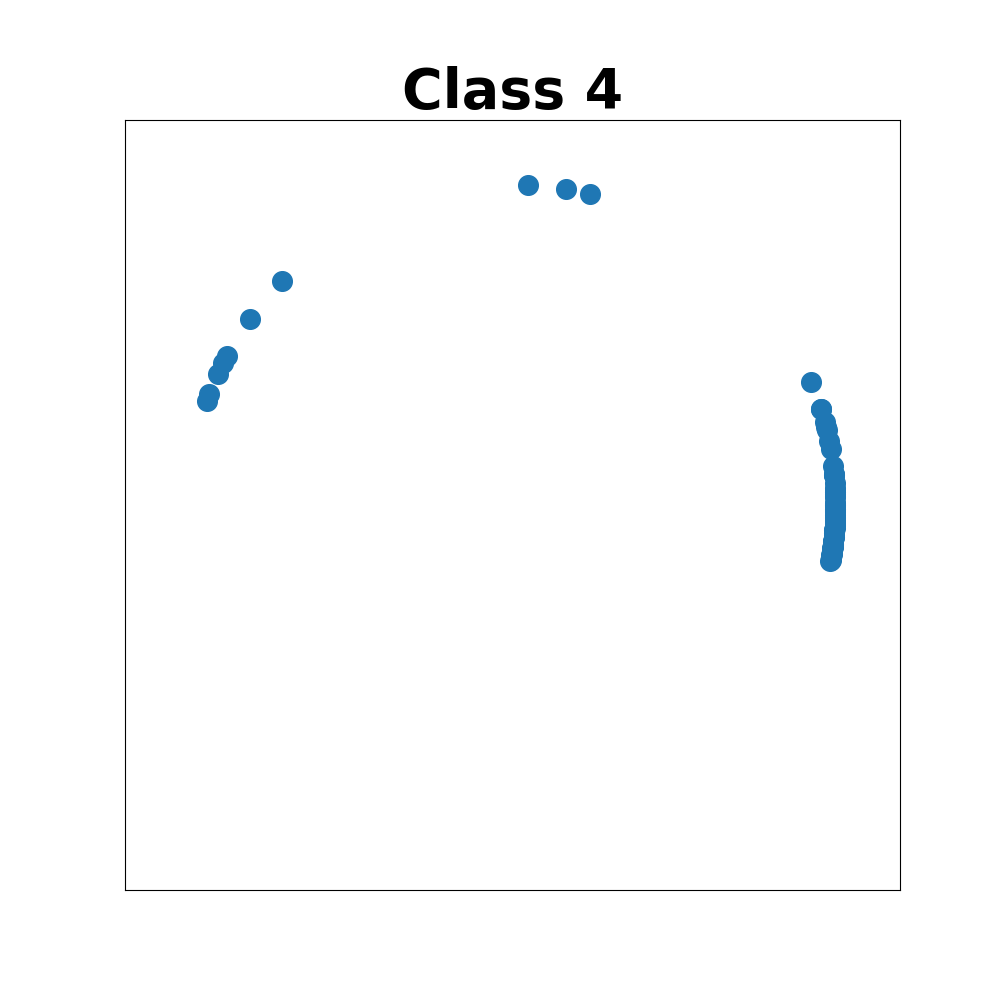}}
	\end{minipage}
	\caption{\label{fig:unif} Node representations of the Cora validation set on the two-dimensional unit hypersphere using CLNR (top) and GRACE \citep{zhu2020GRACE} (bottom). The dashed line in the leftmost plot represents the mean alignment metric. Note the significant improvement in uniformity exhibited by CLNR compared to GRACE.}
\end{figure}
\\
\\
\xhdr{Batch Normalization to Enhance Uniformity} \cite{wang2020understanding} empirically shows that better representations achieve lower value of alignment and uniformity metrics.  Leveraging  these insights, we explain the success of our batch normalization by its ability to enhance the uniformity of the representation. Indeed, batch normalization has the effect of scaling all dimensions of the embedding space --- thereby making all embedding directions comparable and the embedding space, isotropic. Such standardization is in fact routine procedure for other unsupervised learning techniques, including principal component analysis, to prevent one dimension of the data from overpowering the others \citep{hastie2009elements}. Similarly here, we posit that BN stabilizes the contribution of each dimension and allows for an easier optimization of uniformity. {\it  We posit that this batch normalization is sufficient to encourage the uniformity that \cite{wang2020understanding} showed to be beneficial for embedding quality, and has the significant advantage of not requiring hyperparameter tuning.}
\\
\\
To provide empirical evidence for this phenomenon and clearly see the effect of each postprocessing, we additionally consider the following variants:
\\
\begin{description}[noitemsep, leftmargin=1.5em, topsep=0em]
	\item[nCLNR:] We first compare the CLNR and GRACE to a counterpart (which we call \textit{nCLNR}) that uses the identity mapping as postprocessing (i.e., no postprocessing is used). This will serve as our ablation benchmark to verify the importance of the postprocessing step.
	\item[dCLNR:] Building off of the rich literature on the topic in DNNs, another choice of column-wise postprocessing  consists in using decorrelated batch normalization (dBN) \citep{huang2018decorrelated}. dBN expands on the previous standardization technique, and further enforces features in different dimension to be uncorrelated. This holds the promise of further allowing a better conditioning the Hessian, but in standard literature, is often considered too costly to implement. To enrich our discussion, we also consider dBN postprocessing step, and refer to the contrastive learning framework with this choice of postprocessing as \textit{dCLNR}.
	\item[GCLNR:] This last variant corresponds to a postprocessing that uses a 2-layers MLPs with nonlinear activation function followed by column-wise standardization (i.e., both row-wise and column-wise postprocessing are used). 
\end{description}
\begin{figure}[t]
	\begin{minipage}[h]{.33\linewidth}
		\centerline{\includegraphics[scale=0.27]{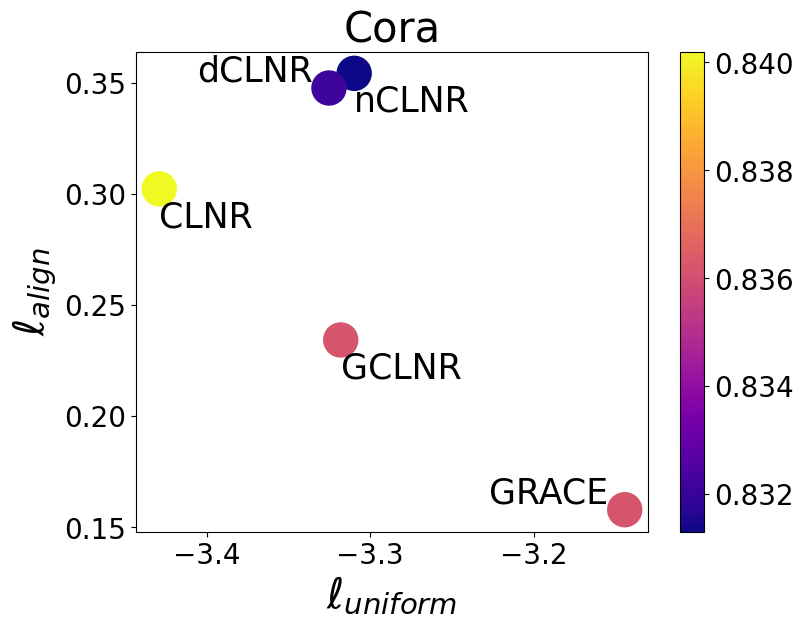}}
	\end{minipage}
	\begin{minipage}[h]{.33\linewidth}
		\centerline{\includegraphics[scale=0.27]{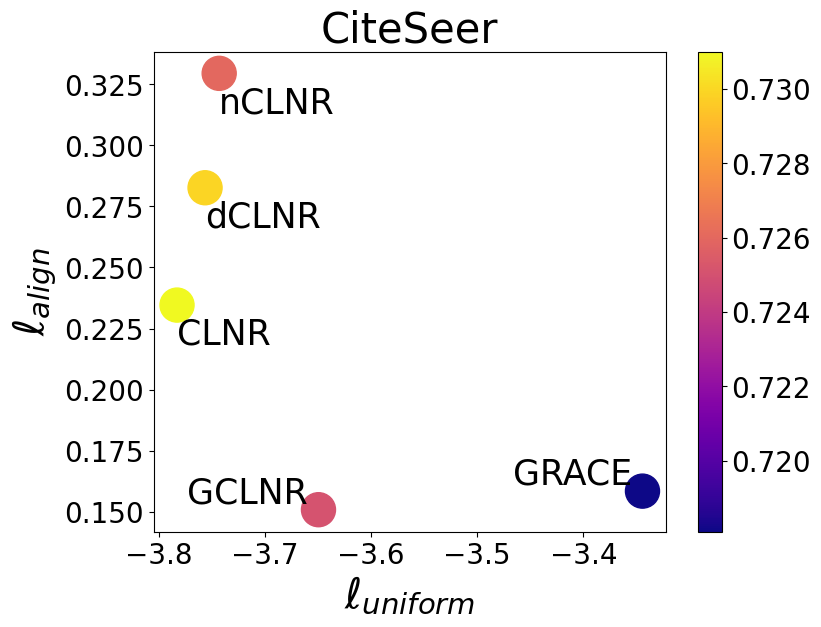}}
	\end{minipage}
	\begin{minipage}[h]{.33\linewidth}
		\centerline{\includegraphics[scale=0.27]{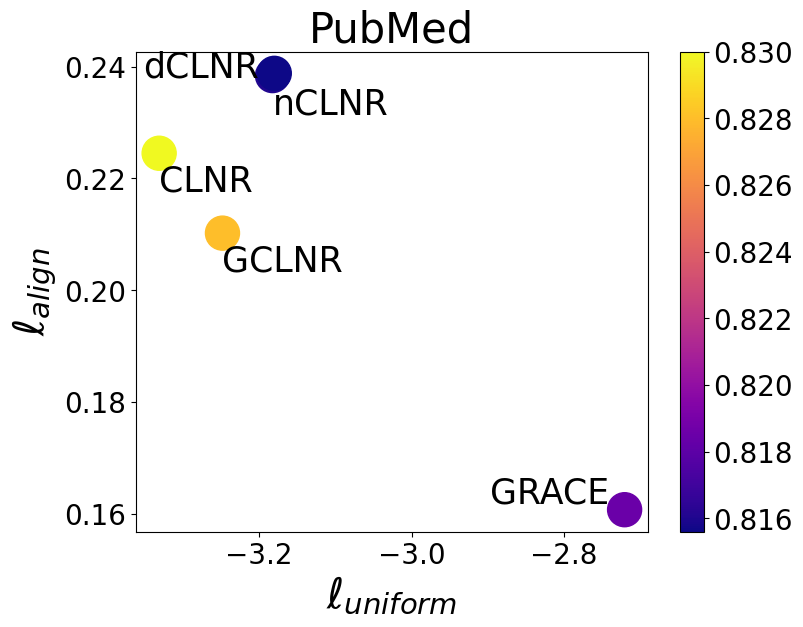}}
	\end{minipage}\\	
	\begin{minipage}[h]{.33\linewidth}
		\centerline{\includegraphics[scale=0.27]{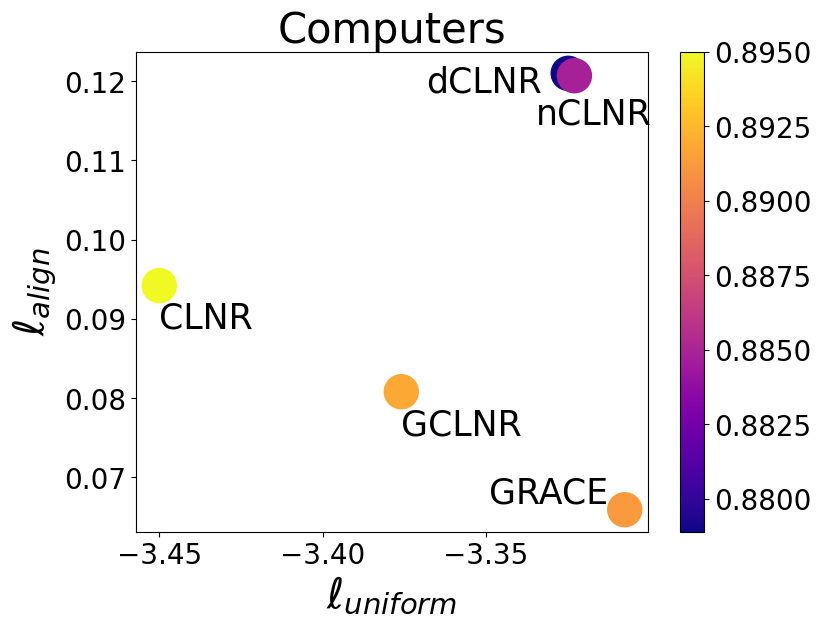}}
	\end{minipage}
	\begin{minipage}[h]{.33\linewidth}
		\centerline{\includegraphics[scale=0.27]{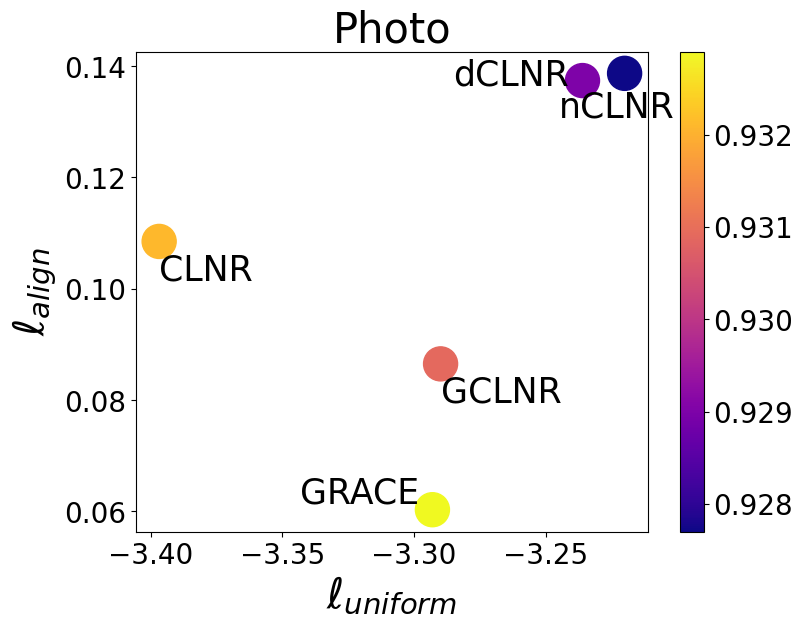}}
	\end{minipage}
	\begin{minipage}[h]{.33\linewidth}
		\centerline{\includegraphics[scale=0.27]{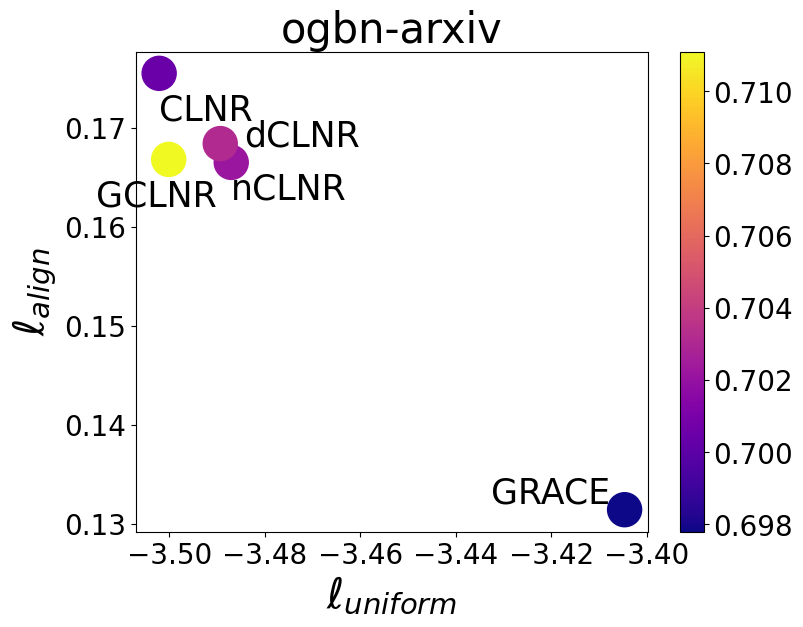}}
	\end{minipage}
	\caption{\label{fig:align_unif} Comparisons between nCLNR, CLNR, GRACE, and their variants dCLNR and GCLNR in terms of their respective alignment and uniformity measures (Equations~\eqref{eq:align} and \eqref{eq:unif}). Each dot represents a method, and while the colour gradient captures the corresponding node classification accuracy.}
\end{figure}
\vspace*{12pt}
We begin by plotting the learned representations on Cora dataset using GRACE and our method, CLNR in Figure~\ref{fig:unif}. As expected, our batch-normalization-based method significantly improved the uniformity of the learned representations.
\\
\\
Figure~\ref{fig:align_unif} shows the alignment (Equation~\eqref{eq:align}) vs uniformity (Equation~\eqref{eq:unif}) metrics achieved for each of the methods on 6 different datasets. We observe that sole CLNR is located in left-bottom side of nCLNR (no postprocessing) for 5 out of 6 benchmark datasets. This implies that column-wise postprocessing improves both alignment and uniformity compared to the one without postprocessing. On the other hand, GRACE is shown consistently at the right-bottom corner of the figures. This implies that row-wise postprocessing focuses on improving alignment of representations at the expense of losing significant amount of uniformity. The only exception is \textit{Photo} where GRACE is also located in left-bottom side of nCLNR and exhibits better node classification accuracy compared to CLNR. We might expect that GCLNR which uses both row-wise and column-wise postprocessing improve the quality of representations compared with the ones using just one of them. Interestingly, we observe that GCLNR is consistently located at the middle of CLNR and GRACE which implies that we can not fully recover the uniformity by column-wise postprocessing once the row-wise postprocessing is done. Surprisingly, dCLNR which additionally decorrelates different features has a similar value of alignment and uniformity metrics
with nCLNR.

\begin{figure}[t]
	\begin{minipage}[h]{.33\linewidth}
		\centerline{\includegraphics[scale=0.27]{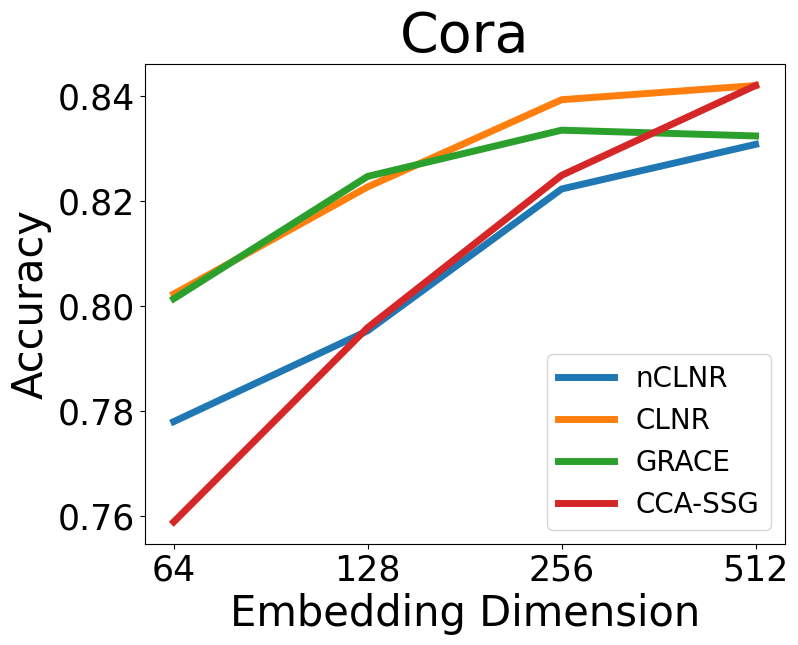}}
	\end{minipage}
	\begin{minipage}[h]{.33\linewidth}
		\centerline{\includegraphics[scale=0.27]{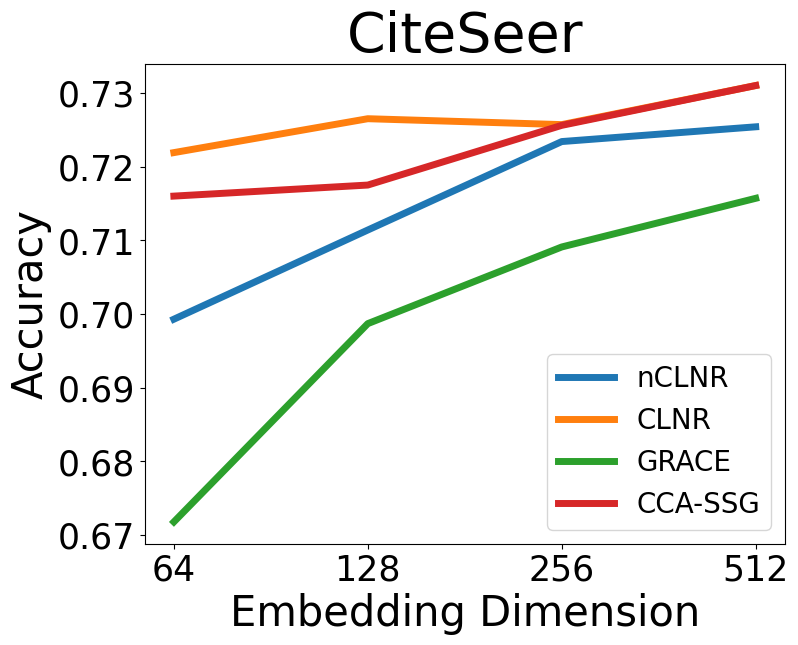}}
	\end{minipage}
	\begin{minipage}[h]{.33\linewidth}
		\centerline{\includegraphics[scale=0.27]{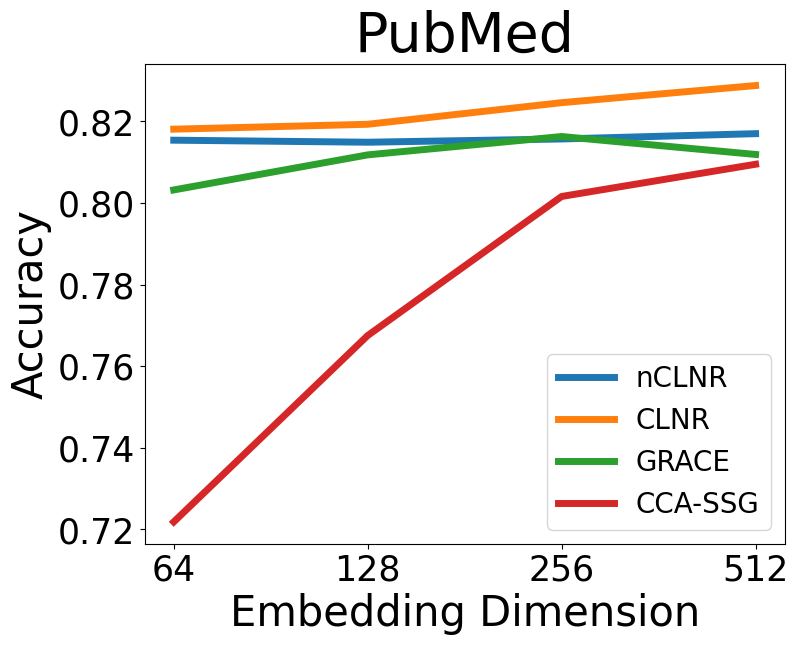}}
	\end{minipage}
	\\
	\begin{center}
		\begin{minipage}[h]{.33\linewidth}
			\centerline{\includegraphics[scale=0.27]{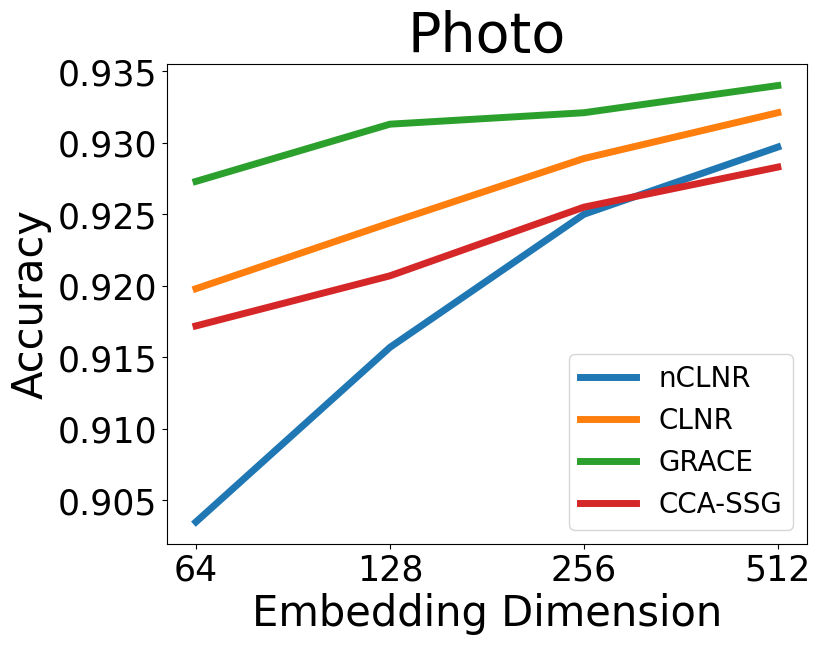}}
		\end{minipage}
		\begin{minipage}[h]{.33\linewidth}
			\centerline{\includegraphics[scale=0.27]{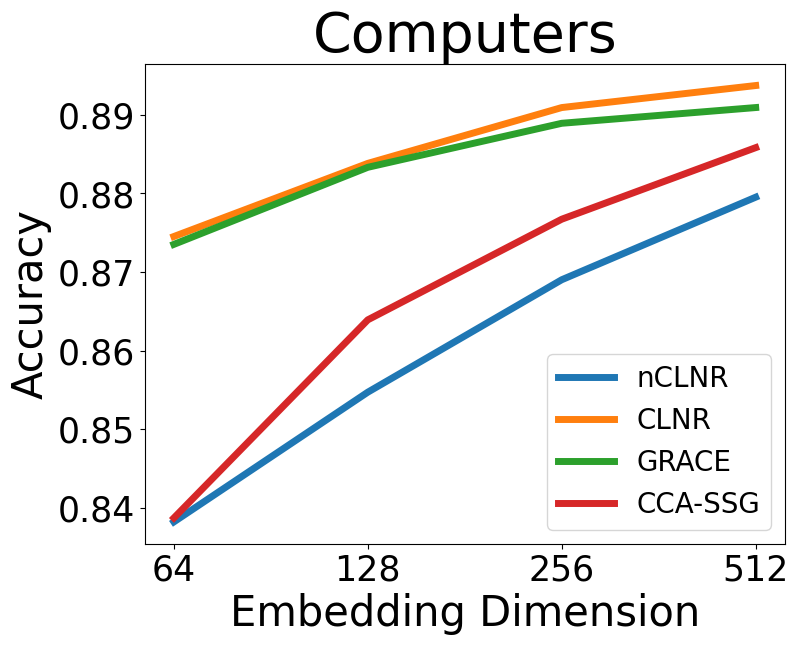}}
		\end{minipage}
	\end{center}
	\caption{\label{fig:embeddings}Comparison between nCLNR, CLNR, GRACE, CCA-SSG as a function of the embedding dimension.}
\end{figure}

\begin{figure}[t]
	\begin{minipage}[h]{.33\linewidth}
		\centerline{\includegraphics[scale=0.27]{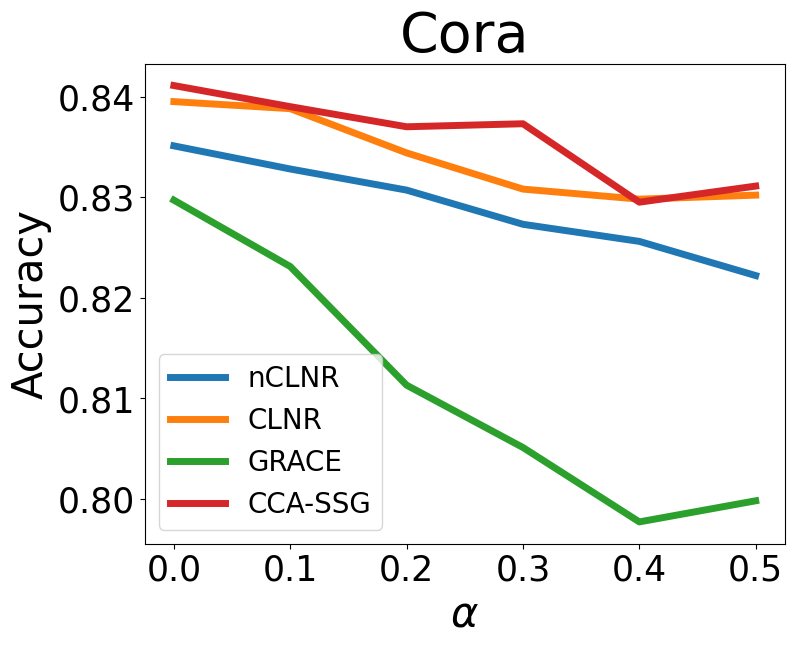}}
	\end{minipage}
	\begin{minipage}[h]{.33\linewidth}
		\centerline{\includegraphics[scale=0.27]{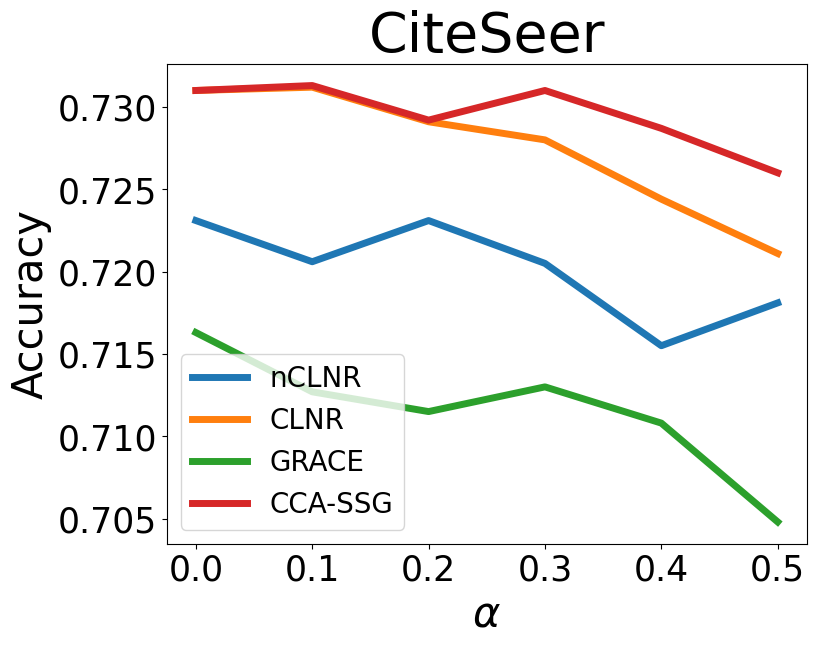}}
	\end{minipage}
	\begin{minipage}[h]{.33\linewidth}
		\centerline{\includegraphics[scale=0.27]{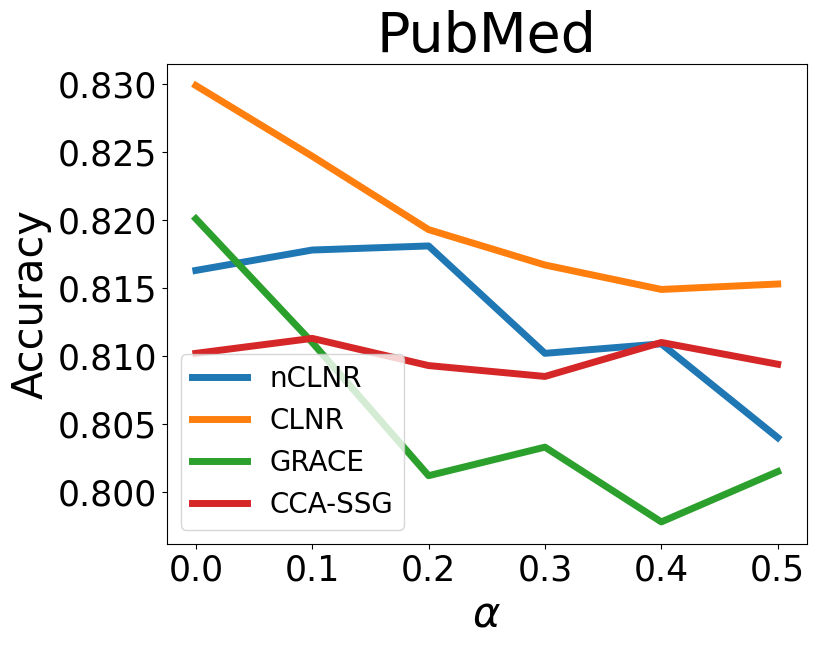}}
	\end{minipage}
	\\
	\begin{center}
		\begin{minipage}[h]{.33\linewidth}
			\centerline{\includegraphics[scale=0.27]{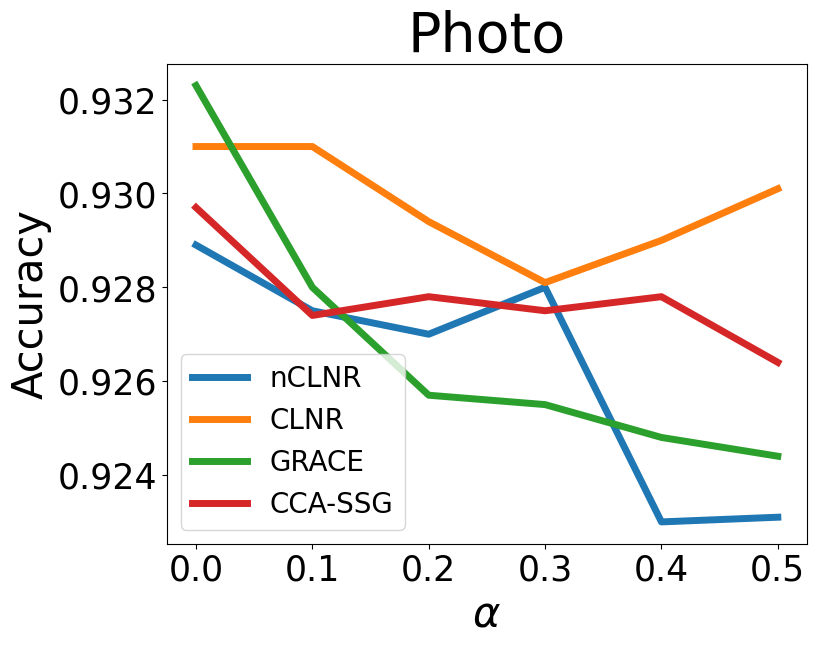}}
		\end{minipage}
		\begin{minipage}[h]{.33\linewidth}
			\centerline{\includegraphics[scale=0.27]{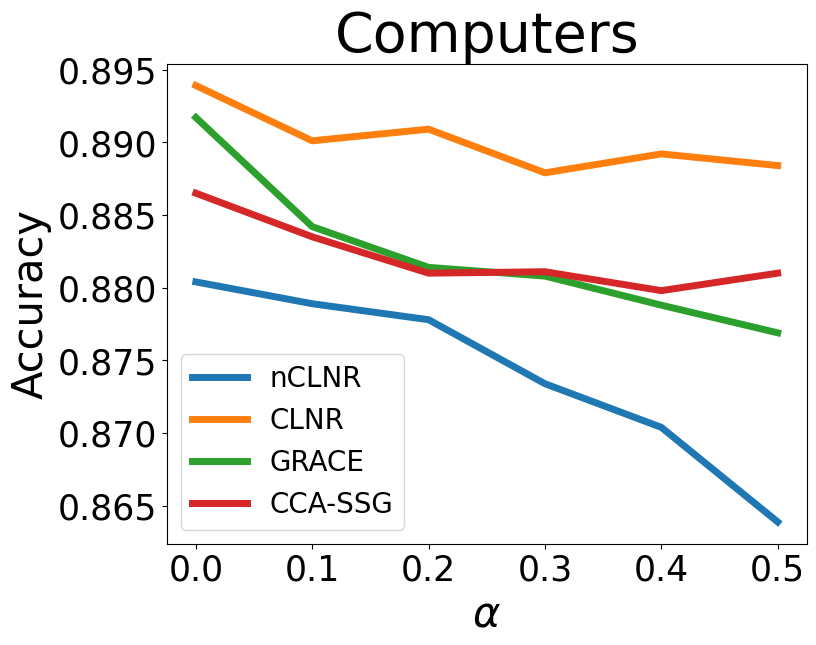}}
		\end{minipage}
	\end{center}
	\caption{\label{fig:robust} Comparison between nCLNR, CLNR, GRACE, CCA-SSG in terms of robustness to edge perturbation.}
\end{figure}

\section{Additional Experiments}\label{sec:sec4}

We conclude this paper with a discussion on the additional benefits of CLNR compared to existing approaches. We focus this discussion on two points: (a) efficiency of the embedding method (as measured by the accuracy of downstream node classifiers as a function of the embedding size) and (b) robustness to perturbations. Throughout this discussion, we include in the comparison a non-contrastive competitor, CCA-SSG \citep{zhang2021CCA-SSG} due to its impressive performance and scalability.
\\
\\
\xhdr{Efficiency of Embedding} We now study the effect of embedding dimension on the quality of node representations. 
From Figure~\ref{fig:embeddings}, we observe that CLNR shows consistent competitive performance across all embedding dimensions on all benchmark datasets. GRACE also shows good performance across all embedding dimensions on all benchmark datasets except \textit{CiteSeer}. Conversely, we observe that nCLNR which does not postprocess the raw embeddings and CCA-SSG which uses a non-contrastive loss are considerably more sensitive to the dimension.
\\
\\
\xhdr{Robustness of Methods}
Finally, we compare nCLNR, CLNR, GRACE, and CCA-SSG in terms of robustness on edge perturbation. Let $\vert A\vert$ be the number of edges in an input graph $G$. For given $p\in(0,1)$ we generate adjacency matrix noise $\Delta_{A}\in\sR^{N\times N}$ containing $\vert A\vert\times p$ number of random undirected edges. We then generate a perturbed adjacency matrix with $\widetilde{A}=A+\Delta_{A}$ and we train GCN encoder with the perturbed graph $\widetilde{G}=(X,\widetilde{A})$ instead of $G$. The results are summarized in Figure \ref{fig:robust}. Figure \ref{fig:robust} shows node classification accuracy of the four models on 5 benchmark datasets along with an increasing value of $p$. We observe that CLNR consistently outperforms all other methods on all benchmark datasets except \textit{CiteSeer}.

\section{Conclusion}\label{sec:sec5}

We have introduced self-supervised contrastive learning for node representations (CLNR), a simplified but powerful method for learning effective node representations. By replacing the expensive MLP-based row-wise postprocessing of GRACE \citep{zhu2020GRACE} with a simple batch normalization, we simplify the architecture of existing contrastive learning methods based on NT-Xent loss and considerably its performance, scalability and robustness. To explain this success, we revisit the alignment and uniformity paradigm. We provided empirical evidence that the column-wise postprocessing was sufficient to improve both alignment and uniformity of the embeddings, while row-wise postprocessing improves the alignment of node representations at the expense of  losing significant amounts of uniformity.

\bibliographystyle{my-plainnat}
\bibliography{ref}

\newpage

\appendix

\section{IMPLEMENTATION DETAILS}

\subsection{Datasets}\label{appen:A1}

We consider eight node-classification benchmarks to evaluate our models: \textit{Cora, CiteSeer, PubMed, Amazon Computers, Amazon Photo, Coauthor CS, Coauthor Physics, ogbn-arxiv}. For all experiments, we use the processed version of the datasets provided by Pytorch Geometric. We provide brief description of the datasets as follows:\\
\\
\xhdr{Cora, CiteSeer, PubMed} A citation network where nodes represent scientific papers and edges mean citations between pairs of paper \citep{sen2008collective}.
\\
\\
\xhdr{Amazon Computers, Amazon Photo} The Amazon co-purchase network where nodes represent goods and edges mean being pairs of goods frequently purchased together \citep{mcauley2015image}. 
\\
\\
\xhdr{Coauthor CS, Coauthor Physics} The Microsoft Academic network where nodes represent authors and edges mean authors who have co-authored a paper \citep{sinha2015overview}.
\\
\\
\xhdr{Ogbn-arxiv} The citation network between all Computer Science (CS) ARXIV papers indexed by MAG where nodes represent ARXIV papers and directed edges mean one paper cites another one \citep{Wang2020Microsoft,hu2020OGB}.
\\
\\
Further details of the datasets are presented in Table~\ref{tab:dataset}.

\begin{table}[h]
	\caption{Dataset Statistics.}\label{tab:dataset} 
	\begin{center}
			\begin{small}
			\begin{sc}
		\begin{tabular}{lcccc} 
		\toprule
			Dataset & $\#$ Nodes & $\#$ Edges & Feature dimension & $\#$ Classes \\ 
			\midrule
			Cora & 2,708 & 10,556 & 1,433 & 7 \\
			CiteSeer & 3,327 & 9,104 & 3,703 & 6 \\
			PubMed & 19,717 & 88,648 & 500 & 3 \\
			Computers & 13,752 & 491,722 & 767 & 10 \\
			Photo & 7,650 & 238,162 & 745 & 8 \\
			CS & 18,333 & 163,788 & 6,805 & 15 \\
			Physics & 34,493 & 495,924 & 8,415 & 5 \\
			ogbn-arxiv & 169,343 & 1,166,243 & 128 & 40 \\
	\bottomrule
		\end{tabular}    
			\end{sc}
\end{small}
	\end{center}
\end{table}

\subsection{Hyper-parameters}\label{appen:A2}

For all datasets, all hyper-parameters are choosen by small grid search. In particular, we fix the temperature parameter $\tau=0.5$ and $\text{wd1}=0$ due to large search space. We provide all details of hyper-parameters that are used in our experiments on the eight benchmarks in Table~\ref{tab:hyperparameter}.

\begin{table}[h]
	\caption{\label{tab:hyperparameter} Details of Hyper-parameters.}
	\begin{center}
			\begin{small}
			\begin{sc}
		\begin{tabular}{lcccccccccc} 
		\toprule
			Dataset & Epoch & $\#$ layers & embedding dimension & $\tau$ & lr1 & wd1 & $p_f$ & $p_e$ & lr2 & wd2 \\ [0.5ex] 
			\midrule
			Cora & 50 & 2 & 512 & 0.5 & 1e-3 & 0 & 0.2 & 0.5 & 5e-3 & 1e-4 \\
			CiteSeer & 50 & 1 & 512 & 0.5 & 1e-3 & 0 & 0.2 & 0.5 & 1e-2 & 1e-2  \\
			PubMed & 600 & 2 & 512 & 0.5 & 1e-3 & 0 & 0.3 & 0.5 & 1e-2 & 1e-4  \\
			Computers & 200 & 2 & 512 & 0.5 & 1e-3 & 0 & 0.0 & 0.5 & 1e-2 & 1e-4  \\
			Photo & 100 & 2 & 512 & 0.5 & 1e-3 & 0 & 0.0 & 0.5 & 1e-2 & 1e-4  \\
			CS & 600 & 2 & 512 & 0.5 & 1e-3 & 0 & 0.3 & 0.1 & 5e-3 & 1e-2  \\
			Physics & 50 & 2 & 512 & 0.5 & 1e-3 & 0 & 0.2 & 0.4 & 5e-3 & 1e-4  \\
			ogbn-arxiv & 5000 & 3 & 512 & 0.5 & 1e-2 & 0 & 0.0 & 0.5 & 1e-2 & 1e-4  \\
	\bottomrule
		\end{tabular}
			\end{sc}
\end{small}
	\end{center}
\end{table}

\section{FURTHER NODE EMBEDDING COMPARISONS}

\subsection{Clustering Score}

We further evaluate the quality of node embedding by examining three clustering scores: \textit{Silhouette Coefficient, Davies-Bouldin index, Calinski-Harabasz score}~\citep{rousseeuw1987silhouettes,davies1979davies,calinski1974calinski}. We consider three methods: CLNR, GRACE, CCA-SSG. We provide brief description of three clustering scores as follows:
\\
\\
\xhdr{Silhouette coefficient} Let us denote Silhouette coefficient for $i$-th node as $SC_i$ and Silhouette coefficient over all nodes as $SC=\dfrac{1}{N}\sum\limits_{i=1}^N SC_i$ where $N$ is the number of nodes. For each node $i$, Silhouette coefficient is calculated by $SC_i =\dfrac{Y_i-X_i}{max(X_i,Y_i)}$ where $X_i$ is the mean intra-cluster distance and $Y_i$ is the mean nearest-cluster distance that $i$-th node is not a part of. The best and worst value that $SC$ attains are 1 and -1 respectively. $SC\approx 0$ indicates overlapping clusters.
\\
\\
\xhdr{Davies-Boulding index} Let us denote Davies-Boulding index as $DB$. Davies-Boulding index is calculated by $DB=\dfrac{1}{K}\sum\limits_{i=1}^K \underset{j\neq i}{\max} R_{i,j}$ where $K$ is the number of classes and $R_{i,j}=\dfrac{S_i+S_j}{M_{i,j}}$ where $S_i$ is the average distance between each node and the centroid of $i$-th cluster, and $M_{i,j}$ is the distance between the centroids of $i$-th and $j$-th cluster. Small value of $DB$ indicates well-separated clusters.
\\
\\
\xhdr{Calinski-Harabasz score} Let us denote Calinski-Harabasz score as $CH$. Calinski-Harabasz score is calculated by $CH=\dfrac{tr(B)}{tr(W)}\times\dfrac{N-K}{K-1}$ where $tr(B)$ is the trace of the between group dispersion matrix, $tr(W)$ is the trace of the within-cluster dispersion matrix, $K$ is the number of classes, and $N$ is the number of nodes. Higher score indicates dense and well-separated clusters.
\\
\\
Clustering scores for CLNR, GRACE, and CCA-SSG on Cora dataset are summarized in Table~\ref{tab:cluster}. We observe that CLNR achieves the best Silhouette coefficient and CCA-SSG achieves the best Davies-Boulding index and Calinski-Harabasz score. Moreover, all clustering scores for CLNR are higher than those of GRACE.

\begin{table}[h]
	\caption{\label{tab:cluster} Clustering Scores on Cora Dataset.}
	\begin{center}
		\begin{small}
			\begin{sc}
				\begin{tabular}{lccc} 
					\toprule
					Methods & $SC$ & $DB$ & $CH$ \\ 
					\midrule
					CLNR & \textbf{0.105} & 2.248 & 63.066 \\
					GRACE & 0.041 & 2.686 & 49.503  \\
					CCA-SSG & 0.089 & \textbf{2.002}  & \textbf{70.760}\\
					\bottomrule
				\end{tabular}
			\end{sc}
		\end{small}
	\end{center}
\end{table}
\numberwithin{equation}{section}
\numberwithin{theorem}{section}

\end{document}